\documentclass[10pt,twocolumn,letterpaper]{article}

\usepackage{iccv}              
\usepackage{graphicx}
\usepackage{amsmath}
\usepackage{amssymb}
\usepackage{booktabs}
\usepackage{booktabs}
\usepackage{multirow}
\usepackage{comment}
\usepackage{colortbl}
\usepackage{musicography}

\definecolor{cvprblue}{rgb}{0.21,0.49,0.74}
\usepackage[pagebackref,breaklinks,colorlinks,allcolors=cvprblue]{hyperref}

\def\paperID{14692} 
\def\confName{ICCV}
\def\confYear{2025}

\title{MeshMamba: State Space Models \\ for Articulated 3D Mesh Generation and Reconstruction}

\begin{document}        

\makeatletter
\let\@oldmaketitle\@maketitle
\renewcommand{\@maketitle}{\@oldmaketitle
\vspace{-25pt}
\centering
  \includegraphics[width=0.95\textwidth]{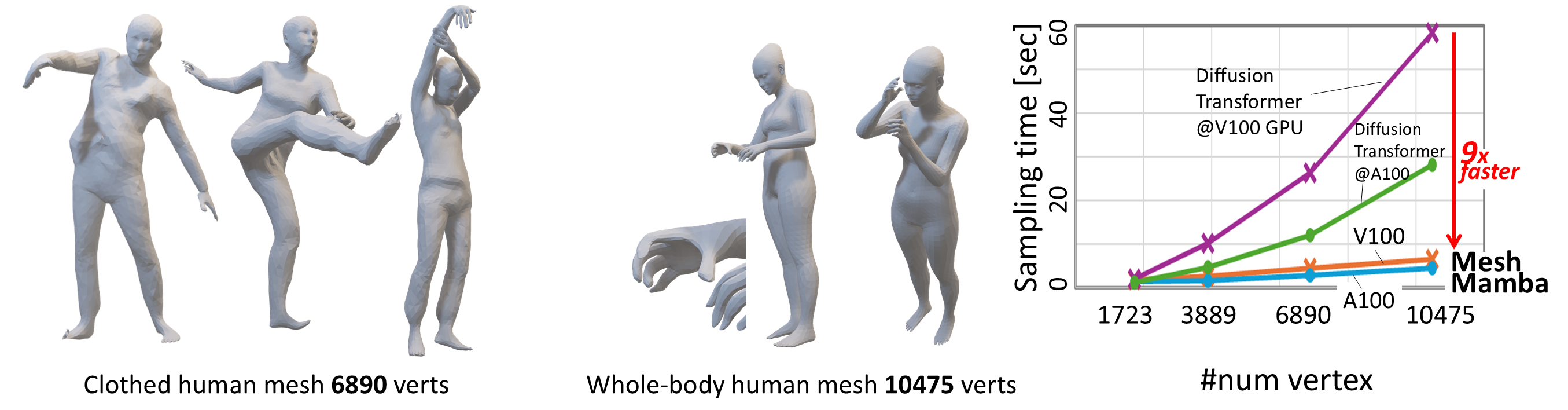}
  \captionof{figure}{Denoising diffusion models based on {\bf MeshMamba} are able to generate dense 3D articulated meshes with around 10,000 vertices, capturing clothing deformations and hand grasp poses. MeshMamba can generate a mesh with 10475 vertices in a few seconds using 100 DDIM sampling steps, which is $6$-$9 \times$ faster than diffusion transformer. 
   \vspace{7pt}}

%
  \label{fig:teaser}
  }
 \vspace{-10mm}

\makeatother

\author{Yusuke Yoshiyasu  \;\;\;\;\;\;\;\; Leyuan Sun \;\;\;\;\;\;\;\; Ryusuke Sagawa\\
 National Institute of Advanced Industrial Science and Technology (AIST)\\
{\tt\small yusuke-yoshiyasu@aist.go.jp}
}

\maketitle

\begin{abstract}

 \vspace{-15pt}
In this paper, we introduce MeshMamba, a neural network model for learning 3D articulated mesh models by employing the recently proposed Mamba State Space Models (Mamba-SSMs). MeshMamba is efficient and scalable in handling a large number of input tokens, enabling the generation and reconstruction of body mesh models with more than 10,000 vertices, capturing clothing and hand geometries. The key to effectively learning MeshMamba is the serialization technique of mesh vertices into orderings that are easily processed by Mamba. This is achieved by sorting the vertices based on body part annotations or the 3D vertex locations of a template mesh, such that the ordering respects the structure of articulated shapes. Based on MeshMamba, we design 1) MambaDiff3D, a denoising diffusion model for generating 3D articulated meshes and 2) Mamba-HMR, a 3D human mesh recovery model that reconstructs a human body shape and pose from a single image.
Experimental results showed that MambaDiff3D can generate dense 3D human meshes in clothes, with grasping hands, etc., and outperforms previous approaches in the 3D human shape generation task. Additionally, Mamba-HMR extends the capabilities of previous non-parametric human mesh recovery approaches, which were limited to handling body-only poses using around 500 vertex tokens, to the whole-body setting with face and hands, while achieving competitive performance in (near) real-time.

\end{abstract}

\section{Introduction}
\label{sec:intro}

Generating and reconstructing 3D articulated mesh models in diverse body shapes and poses is a crucial problem in computer vision and computer graphics, with broad applications in VR, AR, gaming and VFX. The main approaches for solving these tasks can be categorized into parametric and non-parametric vertex-based  paradigms \cite{tian2022hmrsurvey}. Parametric approaches \cite{hmrKanazawa17, pymaf2021,pymafx2022} rely on human body models, such as SMPL \cite{SMPL:2015} and SMPL-X \cite{SMPL-X:2019}, to represent a human body using shape and pose parameters. In contrast, vertex-based approaches \cite{lin2021end-to-end, cho_arxiv.2207.13820} directly manipulate the mesh vertices of a surface and reconstruct them using neural networks. The first paradigm dominates the current field due to its compact representation of body kinematics, while the latter employs a neural network friendly representation of a 3D surface \cite{sarandi2024nlf,POP:ICCV:2021,Zheng2023pointavatar} and holds the potential to capture complex deformations including those of clothing in a general and unified manner. In both paradigms, transformers have become the dominant architecture which offers large improvements in reconstruction performance especially when a large-scale training data is available.

However, the main challenge with transformers is their quadratic complexity with respect to the input sequence length \cite{10.1145/3530811}. In particular, vertex-based transformer approaches are typically limited to processing coarse-resolution meshes with around 500 vertices \cite{lin2021end-to-end,lin2021-mesh-graphormer} due to memory consumption and inference speed constraints. These approaches thus requires an additional upsampling process to obtain a full-resolution mesh with several thousand vertices but it would lose local geometric shapes, which is why the current approaches are limited to body-only pose reconstruction without hand pose and facial expression. 

State Space Models (SSMs) are a family of sequence model that extend RNNs and have recently attracted attention as a potential next-generation sequence model following transformers \cite{gu2022efficiently}. By representing transitions between time frames using a linear system, a sequence can be processed using convolution. Unlike RNNs, SSMs can therefore be trained on all the frames simultaneously, akin to transformer, while still maintaining efficient inference speed. Notably, Mamba \cite{gu2024mamba}  introduced hardware-friendly selective mechanisms for modeling transitions from data, enhancing the expressivity of SSMs. Mamba has quickly spread to various computer vision tasks, including processing images, videos  and 3D point clouds \cite{liang2024pointmamba}. The key to gaining Mamba's potential in these domains lies in how to serialize the input data into a sequence, as opposed to transformer that is agnostic to sequence ordering.

In this paper, we present a method for generating and reconstructing dense 3D articulated mesh models based on Mamba-SSMs, dubbed MeshMamba. To serialize mesh vertices into a sequence that is easier for SSMs to process, we propose a vertex serialization technique that exploits body part UV maps \cite{Guler2018DensePose} and the 3D coordinates of a template body model. Additionally, we explore ways to preserve local geometry of a dense mesh surface by incorporating surface normals through gradient domain mesh representation and a training loss.

Experimental results show that MeshMamba outperforms previous generative models in unconditional 3D human mesh generation tasks. More importantly, MeshMamba can generate dense human body meshes with  more than 10,000 vertices, capturing clothing deformation and hand grasp poses (Fig. \ref{fig:teaser}). Furthermore, we present a novel Mamba-based whole-body 3D human mesh recovery approach, which runs in real-time.

The contributions of this paper includes:
\begin{itemize}
    \item {\bf MeshMamba: } a network model for learning dense 3D articulated meshes based on Mamba-SSMs. We design a serialization technique of mesh vertices based on body part UV maps and the 3D coordinates of a template body mesh for effective training of MeshMamba.  
    
    \item {\bf MambaDiff3D: } a denoising diffusion model for generating 3D articulated meshes based on MeshMamba.  MambaDiff3D is able to generate whole-body human body models, capturing deformations of clothing and hands. It is faster than the transformer-based approach by a factor of $\times 6$-$9$ and outperforms previous generative models in the unconditional 3D human generation task.   

    \item {\bf Mamba-HMR: } a method for 3D human mesh reconstruction from a single image. Mamba-HMR performs competitively with previous approaches in whole-body human mesh reconstruction. It increases the number of input vertex tokens to more than 10,000 vertices, while running at a (near) real-time rate.  
\end{itemize}

\begin{figure*}[t]
\begin{center}
 \includegraphics[width=1.0\linewidth]{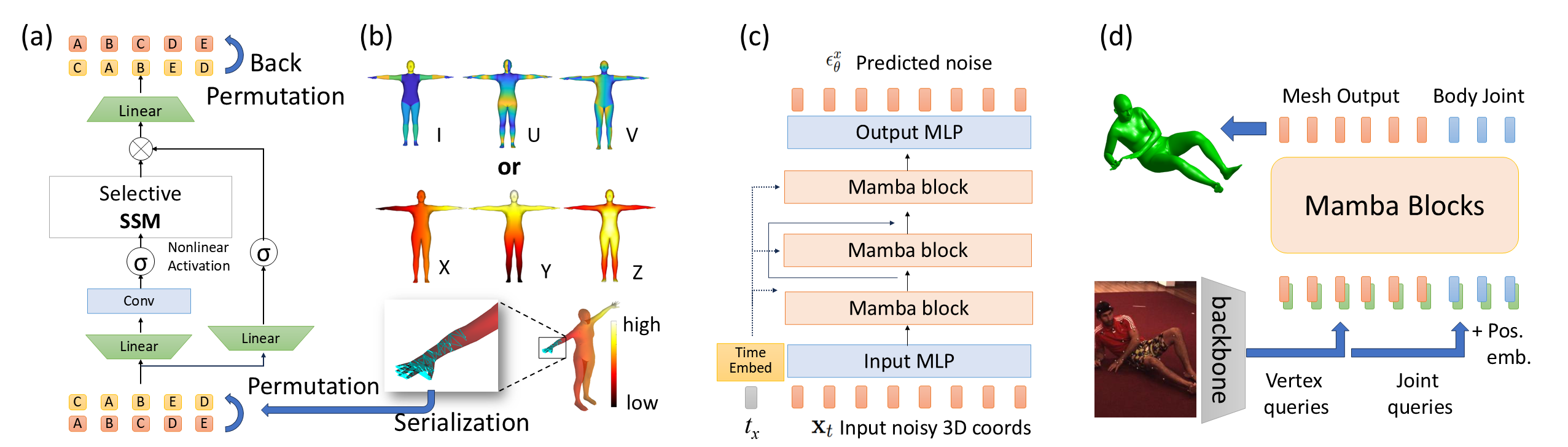}

 \caption{Network block and architectures of MeshMamba. (a) Mamba block with feature permutation based on serialized tokens.  (b) Vertex serialization using DensePose IUV annotations or $xyz$ vertex coordinates of a template mesh. (c) Our diffusion model takes in the noisy 3D coordinates of surface vertices ${\bf x}_t \in  \mathbb{R}^{N \times 3}$ and predicts noise.  (d) Our 3D human mesh recovery model extracts image features from CNN and inputs joint queries and mesh vertex queries to Mamba blocks, along with position embedding. } 

 \label{fig:architecture}
\end{center}
\end{figure*}

\section{Related Work}
\label{sec:related}

\noindent {\bf State space models (SSMs) } Drawing inspiration from the continuous formulation of state space models in control theory, SSMs have been proposed as a solution to efficiently learn long-range dependencies in input sequences \cite{wang2024statespacemodelnewgeneration,patro2024mamba, heidari2024computation}.  Notably, Mamba \cite{gu2024mamba} introduced a selective scan mechanism to enhance the expressiveness of SSMs by modeling transitions between time frames as a function of the input data. Although Mamba was originally proposed for learning long-range sequences in time-series data such as signals and language, it has quickly been adopted across various domains. In the vision domain, ViM \cite{vim} was proposed to enhance vision transformers by employing a bi-directional scan mechanism to handle high-resolution images. DiS \cite{FeiDiS2024} extended the ViM model to the image generation task. Mamba has been adapted to the 3D domain, so far, for point cloud processing and analysis \cite{liang2024pointmamba,zhang2024point}. 
 
\noindent {\bf Generative models for 3D shape and pose } For 3D shape and pose reconstruction, various generative models have been used to build 3D pose and shape priors for various downstream tasks: e.g. variational autoencoders (VAEs) \cite{8578710, yuan2020mesh, hierachical2020, LIMP2020, SMPL-X:2019,COMA:ECCV18, ma2020cape, Jiang2020HumanBody, 9010824, zhou20unsupervised, SemanticHuman}, generative adversarial networks (GANs) \cite{hmrKanazawa17, 9879900, DBLP:journals/corr/abs-1903-10384,iepgan2021}, normalizing flows  \cite{xu2020ghum, biggs2020multibodies, zanfir2020weakly, kolotouros2021prohmr} and diffusion models \cite{ gong2023diffpose, shan2023diffusion, zhang2024rohm, lu2024dposer}. Some works \cite{huang2021arapregasrigidaspossibleregularization, aigerman2022neural} combine generative models and gradient-domain deformable models to achieve detail-preserving shape deformation with neural network models. Extending the ideas from 2D \cite{ho2020denoising, rombach_2022_CVPR} and 3D domain  \cite{luo2021diffusion, zeng2022lion}, recent works utilize diffusion models in 3D human recognition \cite{gong2023diffpose, shan2023diffusion, 10161247, cho2023generative, li2023diffhand, dat2023}. ScoreHMR effectively solves inverse problems for various applications \cite{stathopoulos2024scoreguided} without retraining the task-agnostic diffusion model by guiding its denoising process with a task specific score. ROHM \cite{zhang2024rohm} and DPoser \cite{lu2024dposer} design human pose and motion priors based on diffusion models.

\noindent {\bf 3D human mesh recovery from image } Human mesh recovery approaches \cite{tian2022hmrsurvey} estimate a 3D human body mesh from a single image or video frames, which can be broadly divided into 1) parametric approaches that regress the body shape and pose parameters of human body models  \cite{bogo2016keep, hmrKanazawa17, pymaf2021,cho2023generative} and 2) non-parametric approaches that learns a regression model from an image to 3D vertex coordinates \cite{kolotouros2019cmr, Choi_2020_ECCV_Pose2Mesh, lin2021end-to-end, yoshiyasu2023-deformer,Ma_2023_CVPR,dat2023}. Transformer has been employed in both parametric and vertex-based human mesh recovery, demonstrating strong performance \cite{lin2021end-to-end, lin2021-mesh-graphormer, cho_arxiv.2207.13820,10096870, osx, goel2023humans}. Building on these transformer network architectures, recent studies have developed foundational models for 3D human body pose and shape reconstruction by learning from various datasets, including both synthetic and real data \cite{smplerx}. For whole-body human mesh recovery that reconstructs not only body pose but also hands and face, the dominant approaches in the field are parametric-based   \cite{Pavlakos2020expose,frankmocap, osx, smplerx, sun2024aios, multi-hmr2024}. Neural localization field (NLF) \cite{sarandi2024nlf} is a recent whole-body human mesh reconstruction technique that rely on a continuous shape representation and is able to learn from different dataset formats e.g body joints, SMPL or SMPL-X meshes. Yet its reconstruction quality around face and hands still has room for improvements.

\section{Background}
\noindent {\bf State space models (SSMs)  } A state-space model represents the dynamics of a system using a set of first-order differential equations which describe linear time-invariant (LTI) systems \cite{gu2022efficiently, gu2024mamba}. A multi-input, multi-output LTI system, where the current inputs and states determine changes in the state space of the system, can be described by the following continuous state-space equation:
\begin{align}
\label{eq:continuous}
h'(t) = {\bf A} h(t) + {\bf B} x(t), \; \; y(t) = {\bf C} h(t)  
\end{align}
where  $x(t)$, $y(t)$ and $h(t)$ are the inputs, outputs and  hidden states of the current system, respectively. ${\bf A} \in \mathbb{R}^{N \times N}$, ${\bf B} \in \mathbb{R}^{N \times 1}$ and ${\bf C} \in \mathbb{R}^{1 \times N}$ are continuous parameter of the system. Based on the zero-order hold (ZOH) rule with a time scale parameter $\Delta$, the continuous state space equation in Eq. (\ref{eq:continuous}) can be discretized as follows:
\begin{align}
h'(t) = \bar{\bf A} h(t) + \bar{\bf B} x(t), \; \; y(t) = \bar{\bf C} h(t) 
\label{eq:discrete}
\end{align}
where $\bar{\bf A}  ={\rm  exp} (\Delta {\bf A})$, $\bar{\bf B}  = (\Delta {\bf A})^{-1} ( {\rm  exp} (\Delta {\bf A}) - I) \cdot \Delta {\bf B}$ and $\bar{\bf C}={\bf C}$ are the discrete parameters. Eq. (\ref{eq:discrete}) can be rewritten using global convolution for parallelization:
\begin{equation}
\bar{\bf  K} = (\bar{\bf C} \bar{\bf B}, \bar{\bf C} \bar{\bf A} \bar{\bf B}, \ldots, \bar{\bf C} \bar{\bf A}^k \bar{\bf B} \ldots) , \; \; y = x * \bar{\bf  K}
\label{eq:conv}   
\end{equation}

\noindent {\bf Mamba} proposes a linear time-variant (LTV) system formulation with a selective scan mechanism by introducing time-varying system parameters. 
\begin{align}
h'(t) = \bar{\bf A(x(t))} h(t) + \bar{\bf B(x(t))} x(t), \; \; y(t) = {\bf C(x(t))} h(t)  \nonumber
\end{align}
This allows Mamba to overcome the limitations of previous SSMs, i.e., the lack of context awareness, in other words their ability to selectively remember or forget relevant information. However, this also makes the convolution computation in Eq. (\ref{eq:conv}) impractical. To address this, Mamba introduces a hardware-aware parallel algorithm for selective scanning, which achieves near-linear complexity.

\section{MeshMamba}

We propose MeshMamba, Mamba-based neural network architectures for 3D mesh generation and reconstruction. To do so, we employ a standard Mamba block \cite{gu2024mamba}, which consists of a selective SSM layer, linear layers, a convolution layer and nonlinear activation layers (Fig. \ref{fig:architecture} (a)). The main challenge in adapting Mamba for 3D data lies in the design of strategy for converting the data into a 1D sequence \cite{liu2024point, zhang2024point,liang2024pointmamba}. We therefore design a serialization technique for mesh vertices (Sec.  \ref{sec:serialize}) based on body parts or a template mesh shape structure (Fig. \ref{fig:architecture} (b)).  With our MeshMamba layer, equipped with this serialization technique, we develop a generative diffusion model for 3D human body mesh generation, named MambaDiff3D (Sec. \ref{sec:diffusion}), and a regression model for recovering a 3D human mesh recovery from an image, named Mamba-HMR (Sec. \ref{sec:hmr}). The network architectures are illustrated in Fig. \ref{fig:architecture} (c) and (d).

\noindent{\bf Notation and assumption } We represent an articulated body using a mesh comprising $N$ vertices and $F$ triangle faces. The 3D positions of vertices are denoted as ${\bf x} \in \mathbb{R}^{N \times 3}$. To train our MeshMamba models, we prepare a template mesh $\mathcal{M}_0$ in a canonical pose, along with training meshes $\mathcal{M}_1 \ldots \mathcal{M}_M$ in various body poses and shapes. We assume that the connectivity of the template and all training meshes is the same; in other words, the training meshes are constructed by fitting the template mesh and the point-to-point correspondences between the meshes are known.

\subsection{Vertex serialization}

\label{sec:serialize}

Unlike transformers, Mamba requires ordered input sequences  \cite{wang2024statespacemodelnewgeneration}. Therefore, it is crucial to design a method for serializing 3D mesh vertices into a 1D sequence  so that Mamba can process them more effectively. Recent Mamba approaches for 3D point cloud analysis use space-filling curves predifined in the volumetric space like z-order and Hilbert curves to sort points \cite{liang2024pointmamba, zhang2024point,liu2024point}. However, these methods are unsuitable for mesh generation and reconstruction tasks, which start from random noise or images and deal with deforming articulated bodies.

Our proposed serialization strategy is as follows. Given training meshes with known correspondences, we can serialize all training meshes consistently using sorting indices derived from a template mesh. We explore two serialization approaches that leverage body part UV maps from DensePose annotations  \cite{Guler2018DensePose} and 3D coordinates of a template body mesh, which derive sorting indices directly from mesh vertices without transforming them into other representations like 3D voxels (Fig. \ref{fig:architecture} (b)). To serialize mesh vertices based on their corresponding 3D coordinates in a template mesh, which is in a T-pose for the human case, we sort the vertices primarily along one of the three axes (e.g., x, y, or z). If the values are identical along the primary axis, we then consider the second axis, followed by the third. Similarly, with the DensePose body part IUV maps, we sort the mesh vertices primarily based on the I segmentation map, followed by the U and V maps.

Combining multiple serialization strategies at each Mamba layer helps MeshMamba learn mesh features effectively as reported in previous works e.g. \cite{zhang2024point}. For T-pose vertex coordinates, we generate six serialization methods by varying the order and sign of the x, y, and z axes: ``xyz", ``-xyz", ``yzx", ``-yzx", ``zxy", and ``-zxy". For DensePose annotations, the 24 body parts are sorted based on their centroid coordinates using the same six variations as with the above template mesh based  serialization. Then, the vertices within each part are sorted by U and V maps. However, changing serialization approaches for all layers requires indexing through tokens or ``gather" operations, which can be time-consuming for a deep model with multiple Mamba layers. To balance computational efficiency and shape reconstruction performance, we found that using a combination of two different serialization strategies is effective. Specifically, one serialization strategy is applied across all Mamba layers except for one layer, where the other strategy is used. %

\subsection{MambaDiff3D }  
\label{sec:diffusion}

\noindent {\bf Network model } Our diffusion model for 3D generation is inspired by U-ViT \cite{bao2022all} and its variants \cite{yoshiyasu2024, FeiDiS2024}, which we name
MambaDiff3D. It takes in the noisy 3D coordinates of surface vertices ${\bf x}_t \in  \mathbb{R}^{N \times 3}$ and predicts noise $\epsilon_\theta^x \in  \mathbb{R}^{N \times 3}$ (Fig. \ref{fig:architecture} (c)). Our MambaDiff3D consists of $L+1$ layers of Mamba blocks and input/output MLP layers. The Mamba blocks are categorized into the first half shallow group with $L/2$ blocks, a mid block and a second half deep group with $L/2$ blocks. Skip connections are used to connect the blocks in the first group to those in the second group. Each Mamba block contains hidden layers with $d$ channels. The input MLP layer converts ${\bf x}_t$ into $d$-dimensional embedding features and the output MLP layer converts the Mamba-processed features into $\epsilon_\theta^x$. The time embedding corresponding to timestep $t_x$ is incorporated to every Mamba block by summation. 

\noindent {\bf Train loss and sampling } We adopt the v-prediction parameterization \cite{salimans2022progressive} for the training objective of MambaDiff3D, along with a cosine variance scheduler. This corresponds to the training loss with the weighting $w_t = e^{-\lambda_t/2}$ \cite{NEURIPS2023_kingma}: 
\begin{equation}
L = \mathbb{E}_{t,{\bf x}_0,\epsilon} \; w_t || \epsilon - \epsilon_\theta({\bf x}_t,t) ||^2_2
\label{eq:loss}
\end{equation}
For sampling, we employ the DDIM \cite{song2020denoising} sampler. We set the diffusion time step to $T =1000$ and tested sampling steps with  [50, 100, 250].

\noindent {\bf Combining surface normals and vertex positions } The recent papers \cite{muralikrishnan2024temporalresidualjacobiansrigfree, yoo2024neuralpose} reported that vertex-based generation is prone to local noise, whereas integrating learned Jacobian-fields \cite{10.1145/1015706.1015736, aigerman2022neural} produces globally distorted meshes likely due to error accumulations in tangential components. As our method generates vertices of a dense mesh, we experience this issue especially when training is not long enough or the number of sampling steps is small. 

Instead of generating 3D positions at vertices or Jacobians at triangles, we perform generation of position and normal at each vertex. Then, inspired by the techniques that transfer details in the gradient domain \cite{Botsch2006DeformationTF, nehab2005, weber2007}, we combine surface normals and positions by solving a Poisson system. This allows for smoother reconstruction by removing noise in vertices, while maintaining surface details and global shape structure (Fig. \ref{fig:representation}). 

Specifically, in a similar manner as in \cite{nehab2005}, the gradient at each triangle $m$ is obtained by combining smoothed vertex positions and surface normals in the gradient domain: ${\bf G}_m = {\bf R}_m {\bf T}_m$, where ${\bf T}_m \in \mathbb{R}^{3 \times 3}$ is the Jacobian of the generated vertices after smoothing and ${\bf R}_m \in \mathbb{R}^{3 \times 3}$ is the relative rotation between the generated normals and those obtained from smoothed vertices. These gradients are then plugged into the Poisson system \cite{10.1145/1015706.1015736, aigerman2022neural} to stitch together into a whole mesh. Note that the right-hand side of the Poisson system does not change for the mesh with the same connectivity. Thus, we can reuse the factorization of the system, thereby maintaining the overall generation time without a large overhead \cite{10.1145/1015706.1015736, aigerman2022neural}. Differently from previous approaches \cite{aigerman2022neural}, our approach is not end-to-end, i.e., the generation and the surface reconstruction by solving the Poisson system are done independently, where no gradient is flowing from the Poisson system to the MambaDiff3D model during training.

\subsection{Mamba-HMR } 
\label{sec:hmr}

We present a simple yet effective vertex-based baseline for human mesh recovery based on our MeshMamba, dubbed Mamba-HMR. 
As Mamba-HMR deals with the full-resolution SMPL and SMPL-X meshes without down sampling them, Mamba-HMR is applicable to both body-only and whole-body settings.

\noindent {\bf Network model } The network architecture of Mamba-HMR follows Mesh transformer \cite{lin2021end-to-end} where we essentially replace their transformer blocks with MeshMamba blocks with our  vertex serialization strategies.  Our Mamba-HMR feeds CNNs image features to Mamba as body joint queries and vertex queries, along with position embedding (Fig. \ref{fig:architecture} (d)). The key difference from previous vertex-based approaches \cite{kolotouros2019cmr,cho_arxiv.2207.13820,PointHMR,dat2023} is that Mamba-HMR does not necessarily need upsamplers and its Mamba-blocks directly output a full-resolution mesh, which leads to a  large reduction in  model parameters. Like our MambaDiff3D, Mamba-HMR consists of the shallow, mid and deep Mamba block groups and uses skip connections, except that we do not input time embeddings.

\noindent {\bf Training loss }  Our training loss follows \cite{lin2021end-to-end,lin2021-mesh-graphormer} but is augmented with local geometric losses such as the surface edge, Laplacian and normal losses, $L_{\rm edge}$, $L_{\rm lap}$ and $L_{\rm normal}$ for regularization.  The total loss is defined as: 
\begin{align}
\nonumber L & = \lambda_{\rm 3D}^{\rm V} L^{\rm V} + \lambda_{\rm 3D}^{\rm J} (L_{\rm 3D}^{\rm J} + L_{\rm reg3D}^{\rm J}) + \lambda_{\rm 2D}^J (L_{\rm 2D}^{\rm J} + L_{\rm reg2D}^{\rm J}) \\ & + \lambda_{\rm edge} L_{\rm edge}  + \lambda_{\rm lap} L_{\rm lap} + \lambda_{\rm normal} L_{\rm normal} 
\end{align}
where $L^{\rm V}$,  $L_{\rm 3D}^{\rm J}$,  $L_{\rm reg3D}^{\rm J}$,  $L_{\rm 2D}^{\rm J}$ and $L_{\rm reg2D}^{\rm J}$ are the vertex, 3D joint, 3D regressed joint, 2D joint and 2D regressed joint loss,  respectively. $\lambda_{\rm 3D}^{\rm V}$, $\lambda_{\rm 3D}^{\rm J}$, $\lambda_{\rm 2D}^{\rm J}$ , $\lambda_{\rm edge}$, $\lambda_{\rm lap}$ and $\lambda_{\rm normal}$ are the weights for controlling the relative strengths of respective terms. 

\begin{figure}[h]
\begin{center}
 \includegraphics[width=1.0\linewidth]{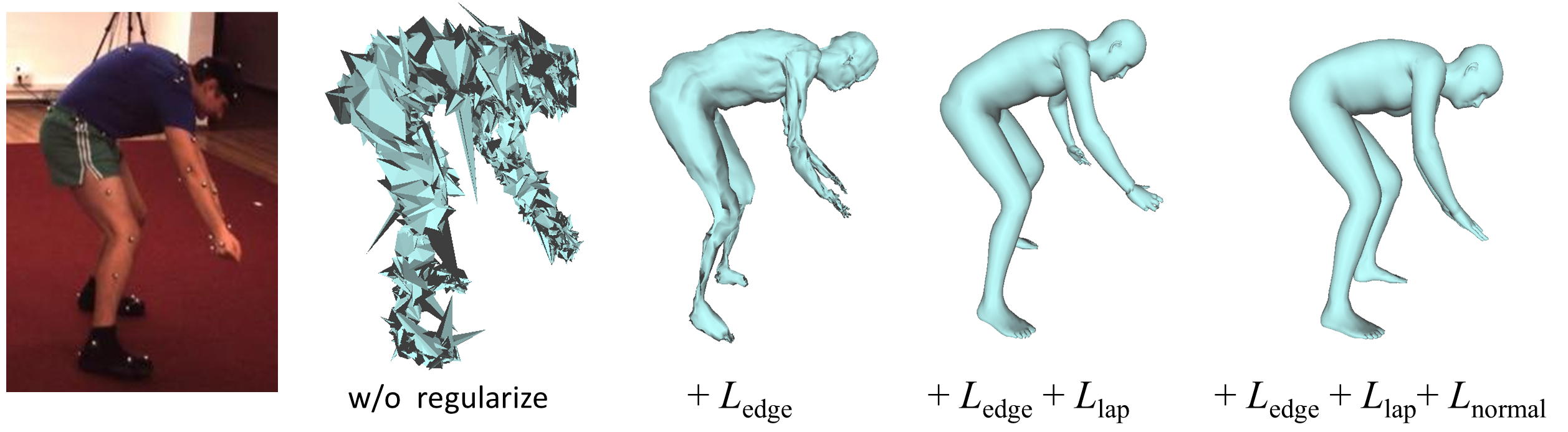} 
 \caption{Importance of local geometric regularization in dense human mesh reconstruction.}
 \label{fig:geoloss}
\end{center}
\end{figure}

The local geometric losses $L_{\rm edge}$, $L_{\rm lap}$ and $L_{\rm normal}$ are vital for local shape preservation in our dense mesh reconstruction (Fig. \ref{fig:geoloss}), which are defined as follows:

\noindent {\bf Laplacian loss }  The  Laplacian loss $L_{\rm lap}$ is written as:
\begin{equation}
L_{\rm lap} = \frac{1}{N} \sum_{i=1}^N || {\bf d}_i -  \bar{\bf d}_i||_1
\end{equation}
where ${\bf d}_i$ and $\bar{\bf d}_i$ are the predicted and ground truth of mean curvature normal vector at vertex $i$ derived from the cotangent Laplacian matrix, respectively.  

\noindent {\bf Edge loss } The edge loss $L_{\rm edge}$ is defined as:
\begin{align}
L_{\rm edge} =  \frac{1}{E} \sum_{e=1}^E  ||{\bf e}_e -  \bar{\bf e}_e||_1
\end{align}
where ${\bf e}_e$ and $\bar{\bf e}_e$ are the predicted and ground truth of edge length  at edge $e$, respectively.  

\noindent {\bf Normal loss } The normal loss $L_{\rm normal}$ is defined as:
\begin{align}
L_{\rm normal} = \frac{1}{F} \sum_{m=1}^F ||{\bf n}_m -  \bar{\bf n}_m||_1
\end{align}
where ${\bf n}_m$ and $\bar{\bf n}_m$ are the predicted and ground truth of face normal at triangle face $m$, respectively.

\begin{figure}[t]
\begin{center}
 \includegraphics[width=1\linewidth]{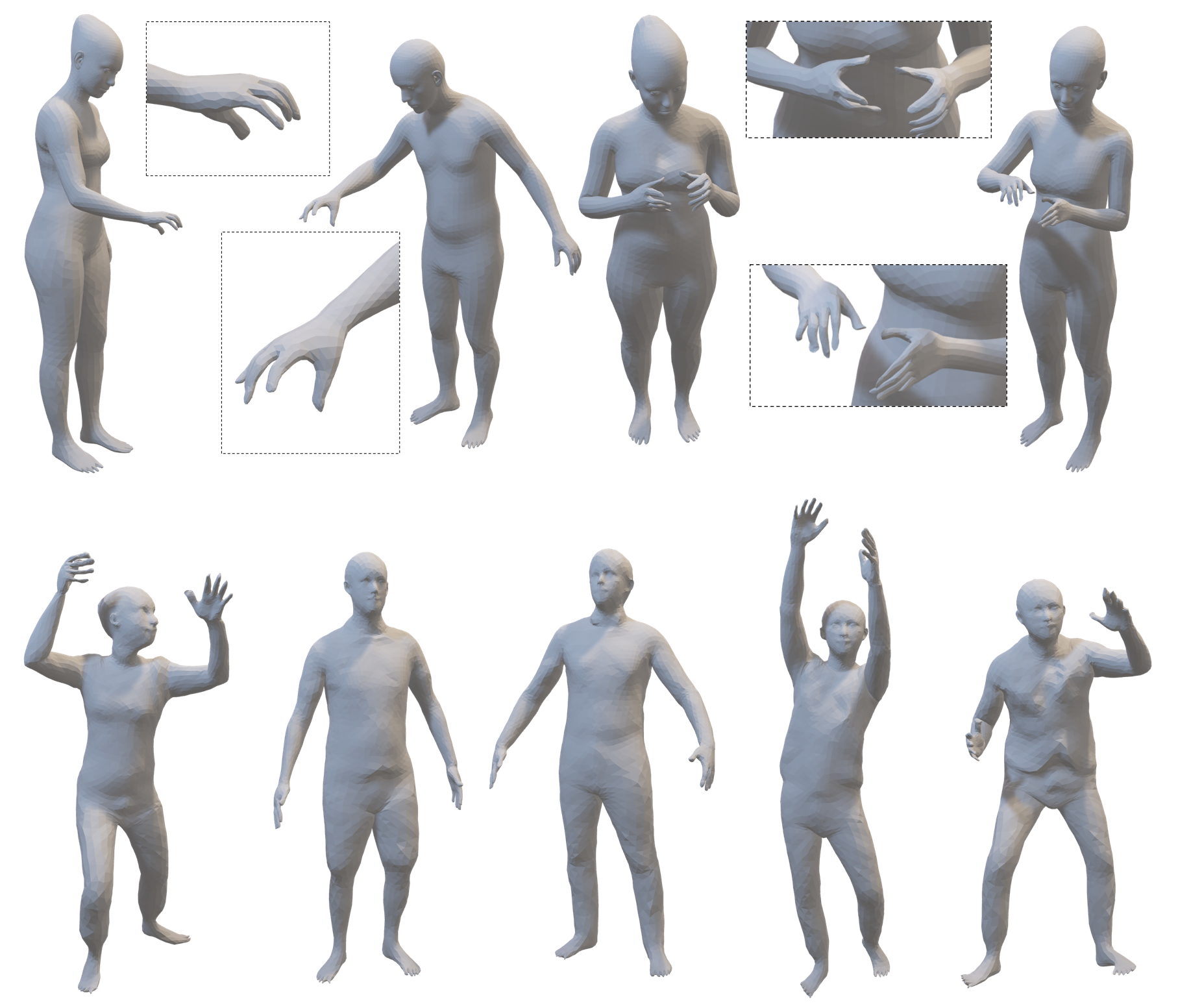}
 \caption{Unconditional generation results of dense 3D meshes. MambaDiff3D can generate human body meshes with 6890 and 10475 vertices, corresponding to the full resolutions of SMPL and SMPL-X, respectively. Notably, MambaDiff3D can capture  grasp hands in GRAB and cloth deformations in CAPE. } 
 \label{fig:dense_mesh}
\end{center}
\end{figure}

\begin{figure}[t]
\begin{center}
\includegraphics[width=1\linewidth]{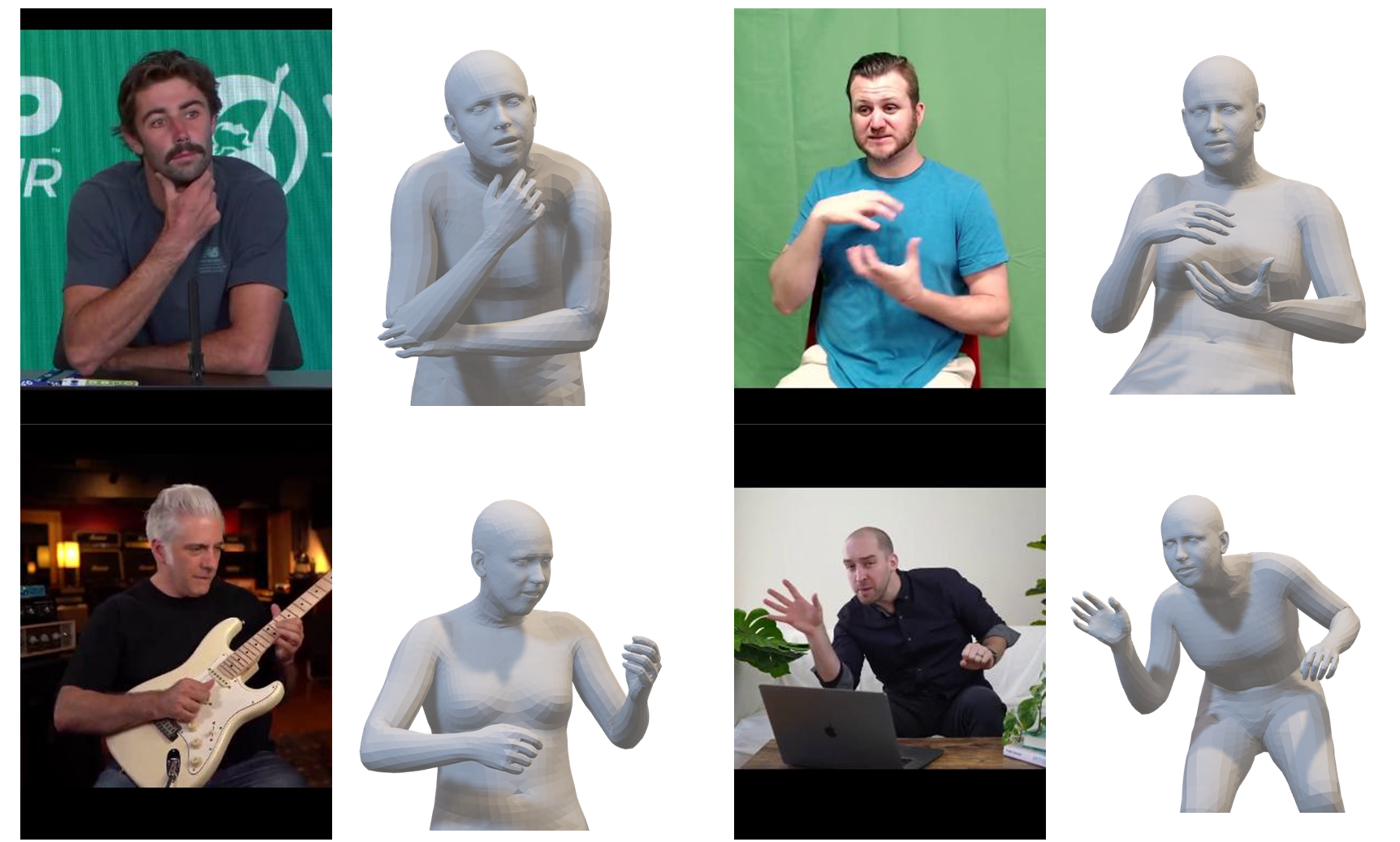}
 \caption{Example results of whole-body 3D human mesh recovery from a single image  using 10475 vertex tokens on UBody.} 
 \label{fig:hmr}
\end{center}
\end{figure}

\begin{figure}[t]
\begin{center}
\includegraphics[width=1\linewidth]{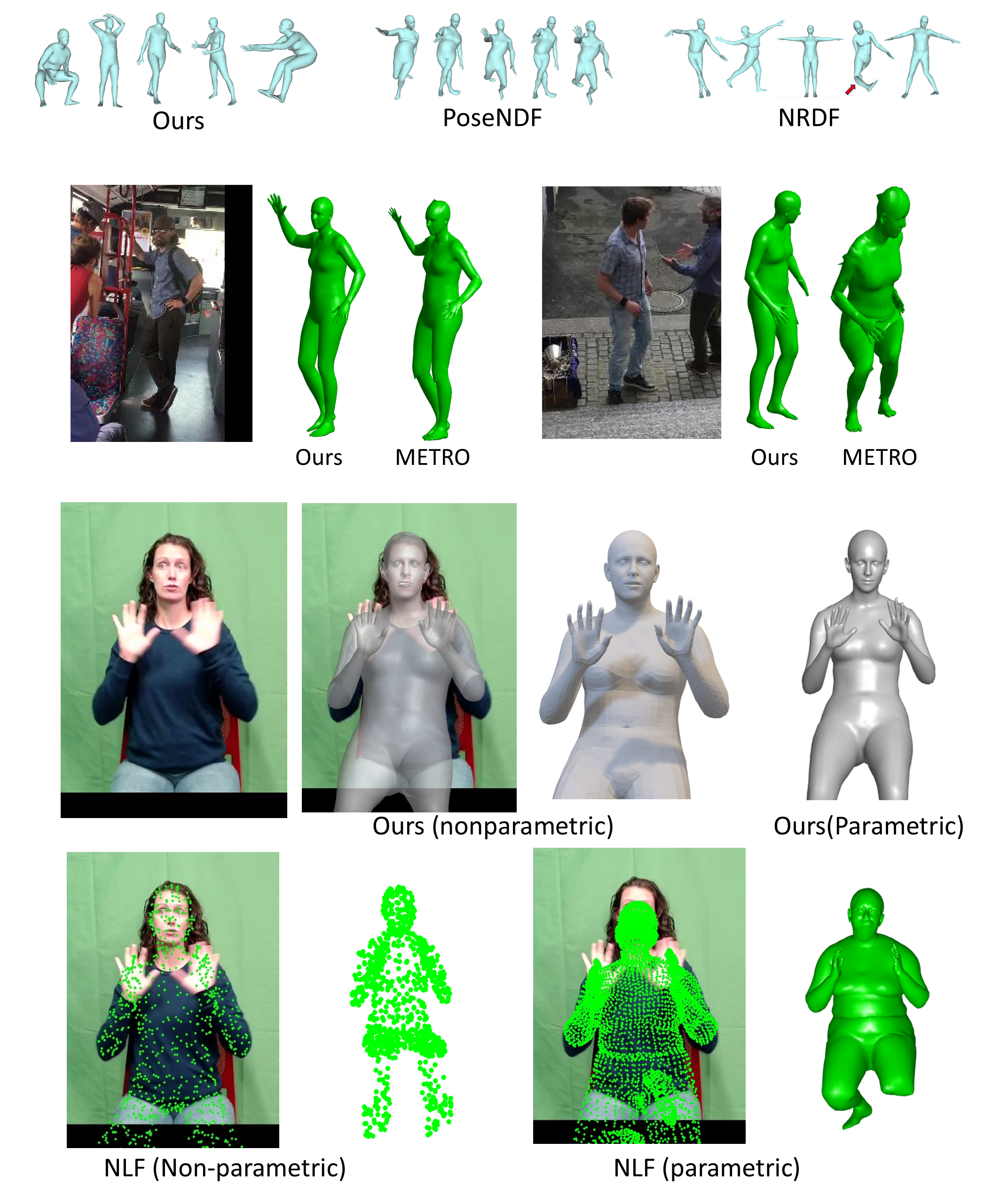}
 \caption{Qualitative comparisons. Top: Human mesh generation VS. DiffSurf, PoseNDF and NRDF. Middle: Body-only 3D human mesh recovery on 3DPW comparing against METRO, where no fine-tuning on 3DPW is performed. Bottom: comparison with NLF on whole-body mesh reconstruction. } 
 \label{fig:comparison}
\end{center}
\end{figure}

\section{Experimental results }

\subsection{Training and evaluation settings}
\subsubsection {\bf 3D articulated mesh generation: MambaDiff3D }

We trained our models on the SURREAL \cite{varol17_surreal}, DFAUST \cite{dfaust2017}, CAPE \cite{ma2020cape}, GRAB \cite{GRAB:2020}, AMASS \cite{AMASS:2019}, BARC \cite{BARC:2022} and Animal3D \cite{xu2023animal3d} datasets. The training meshes were preprocessed to align their global positions and orientations at the root. Our models were trained using either a single cluster node with 8 NVIDIA A100 GPUs or 6 nodes with 4 NVIDIA V100 GPUs. We used the Adam optimizer for training. The learning rate was reduced by a factor of 10 after $1/2$ of the total training epochs beginning from $1 \times 10^{-4}$.

MambaDiff3D is compared against the following pose-based baselines: VPoser \cite{SMPL-X:2019}, Pose-NDF \cite{tiwari22posendf}, NRDF \cite{he24nrdf}, and denoising diffusion on SMPL parameters (Param. Diff.). For these pose approaches, identity parameters are drawn from the standard deviations of the AMASS dataset. We also compared MambaDiff3D with the vertex-based baselines GDVAE \cite{9010824}, LIMP \cite{LIMP2020}, and DiffSurf \cite{yoshiyasu2024}. Evaluation of 3D human generation was conducted on the SURREAL test set (200 meshes). We used the 1-NNA metric \cite{pointflow}, the standard metric in 3D shape generation for quantifying the distributional similarity between generated shapes and the validation set. We also employed the FID and APD metrics used in pose generation \cite{he24nrdf}, which calculates scores from joint locations.

\begin{table}[t]
\centering
\caption{Comparisons with other generative models for unconditional human generation. The 1-NNA metric [\%] assesses the diversity and quality of generated results.  A lower value on this metric signifies superior performance.}
\scalebox{0.8}[0.8]{
\begin{tabular}{c c c c c }
\hline 
Method & Train Set & 1NNA [\%] $\downarrow$  & FID $\downarrow$  & APD $\uparrow$ \\
\hline 
Pose-NDF \cite{tiwari22posendf}  & AMASS & 92.0 & 3.92 & 37.81 \\ 
NRDF \cite{he24nrdf}   & AMASS & 81.6 & 0.64 &  23.12 \\
VPoser \cite{SMPL-X:2019} & AMASS & 60.7 & 0.05  & 14.68
 \\
Param diff & AMASS & 59.6 & --- &  ---\\
GDVAE \cite{9010824} & SURREAL  &  93.8 & --- & --- \\
LIMP \cite{LIMP2020} & FAUST & 81.3  & --- & ---\\
DiffSurf \cite{yoshiyasu2024} & SURREAL & 54.4 & --- & --- \\
Ours (MambaDiff3D) & SURREAL & {\bf 53.1} & 0.32 & 23.01 \\
Ours (MambaDiff3D) & AMASS & 55.1 & 0.22 & 23.8 \\ 
\hline 
\end{tabular}
\label{tab:generation}
}
\end{table}

\begin{table}[t]
\centering
\caption{Comparisons with whole-body 3D mesh recovery approaches on UBody. $\dag$ indicates fine-tuned on UBody. }
\scalebox{0.75}[0.8]{
\begin{tabular}{c c c c c c c c}
\hline 
& \multicolumn{3}{c}{PA-MVE $\downarrow$ (mm)} &\multicolumn{3}{c}{MVE $\downarrow$ (mm)} & \\
Method & All & Hands& Face& All& Hands& Face  & FPS \\
\hline 
OSX-L \cite{osx} & 42.4 &10.8 & 2.4 & 92.4 & 47.7 & 24.9 & 14 \\
OSX-L \cite{osx} $\dag$ &  42.2  & 8.6 & {\bf 2.0} & 81.9 & 41.5 & 21.2 & 14 \\
SMPLer-X-L \cite{smplerx} & 33.2 & 10.6 & 2.8  & 61.5 & 43.3 & 23.1  & 24 \\
SMPLer-X-L \cite{smplerx} $\dag$ & 31.9 & 10.3 & 2.8 & 57.4 & 40.2 & 21.6 & 24  \\
AiOS \cite{sun2024aios} & 32.5& {\bf 7.3} & 2.8& 58.6& 39.0 & 19.6 & --- \\
Multi-HMR-B \cite{multi-hmr2024} & 31.4 & 9.8 & 6.1 & 65.1 & {\bf 33.1} & 22.6 & 23 \\
NLF-L \cite{sarandi2024nlf} & 66.8 & 19.4 & 6.6 &--- &  --- & --- & {\bf 41} \\
\hline
Ours & 26.3 & 10.7 & 2.4 & 54.4 & 38.8 & 17.7 & 22 \\
Ours $\dag$ & {\bf 25.9} & 9.7 & 2.1 & {\bf 51.7} & 33.9 & {\bf 15.9} & 22 \\
\hline 
\end{tabular}
}
\label{tab:hmr-wb2}
\end{table}

\begin{table}[t]
\caption{Ablation studies on network layer blocks and serialization methods. The 1-NNA metric [\%] $\downarrow$ is used. }
\label{tab:ablation}
\vspace{-5pt}
\begin{subtable}{0.225\textwidth}
\centering
\scalebox{0.9}[0.9]{
\begin{tabular}{c c c }
\hline 
Block  & 1NNA $\downarrow$ \\
\hline 
MLP  & 73.7 \\
GNN & 74.2 \\
Transformer  &  53.6 \\
Mamba  & 53.1 \\
\hline 
\end{tabular}
}
 \end{subtable}%
       \begin{subtable}{0.25\textwidth}
       \centering
\scalebox{0.8}[0.8]{
\begin{tabular}{c c c }
\hline 
Serialization &  1NNA $\downarrow$\\
\hline 
SMPL connectivity $\times$ 1 & 60.0 \\
part-IUV $\times$ 1 &  54.4 \\
part-IUV $\times$ 2 &  53.7 \\
part-IUV $\times$ 4 &  53.7 \\
part-IUV $\times$ 7 &  53.0 \\
SMPL $\times$ 1 + IUV $\times$ 1 &  53.5 \\
SMPL $\times$ 1 + XYZ $\times$ 1  &  53.1 \\
SMPL $\times$ 1 + XYZ $\times$ 6  & 53.5  \\
\hline 
\end{tabular}
}
 \end{subtable}%
\vspace{-10pt}
\end{table}

\subsubsection{Human mesh recovery: Mamba-HMR }

We evaluated our method on UBody comparing against the state-of-the art approaches: OSX \cite{osx}, SMPLer-X \cite{smplerx}, AiOS \cite{sun2024aios}, Multi-HMR \cite{multi-hmr2024} and NLF \cite{sarandi2024nlf}, which run faster than interactive rate 10 FPS and are trained on various dataset or fine-tuned on the UBody dataset \cite{osx}. Following previous approaches such as \cite{osx,multi-hmr2024, smplerx}, Mamba-HMR is trained on Human3.6M \cite{h36m_pami},  COCO \cite{LinMBHPRDZ14}, AGORA \cite{agora2021}, BEDLAM \cite{bedlam} and UBody \cite{osx}. We employ HRNet-w48 \cite{sun2019deep} as our CNN backbone, initialized with the weights pre-trained on the 2D human pose detection tasks. It uses a $384 \times 288$ image as input and extracts an $12\times9$ feature map, which is pre-trained on the COCO-whole body dataset \cite{jin2020whole}. The weights in the Mamba blocks are randomly initialized. Whole-body HMR results on EHF and AGORA-val, as well as body-only results on Human3.6M and 3DPW, are provided in the Appendices.

\noindent {\bf Evaluation metrics } We used the following metrics for evaluation. Mean-per-Vertex-Error (MVE) measures the Euclidean distances between the (pseudo) ground truth and the predicted vertices. The PA-MVE metric, where PA stands for Procrustes Analysis, measures the reconstruction error after removing the effects of scale and rotation. All reported
errors are in units of millimeters.

\subsection{Inference and training efficiency}

Figure \ref{fig:teaser} shows a comparison between Mamba and  transformer in terms of inference speeds when used as a layer block in denoising diffusion models. The gap between the two widens as the number of token increases. On an NVIDIA A100 GPU, it takes approx. 4.5 sec for Mamba to generate a mesh with 10475 vertices using 250 DDIM sampling steps, whereas it takes 28.1 sec for the transformer with Pytorch Flash attention enabled. In this case, Mamba is $6 \times$ faster than transformer. Notably, with 50 DDIM steps, MambaDiff3D can generate reasonable quality meshes of the same resolution in about 1 sec. On a V100 GPU where hardware optimization is not available, Mamba is about $9 \times$ faster than transformer (6.6 sec VS. 58.3 sec). These results highlight the scalability of Mamba  w.r.t the number of input tokens. Furthermore, the training time of MeshMamba for 6890 vertex tokens is approx. 18 min per epoch using $6 \times 4$ Nvidia V100 GPUs with batch size of 8, compared to 100 minutes for the transformer under the same settings.

\subsection{Qualitative results} Figures \ref{fig:teaser}  and \ref{fig:dense_mesh} show some example results of unconditional and class conditional 3D human generation. As visualized, MambaDiff3D can generate 3D human meshes in diverse body shapes and poses, including grasping hands and cloth deformations. Given clothing types as conditions, MambaDiff3D can generate human meshes in different clothing styles such as blazer and polo from the CAPE dataset.

Figure \ref{fig:hmr} shows the results of whole-body 3D human mesh recovery using Mamba-HMR on UBody. Mamba-HMR can reconstruct a realistic, dense 3D human mesh from a single image. In Figure \ref{fig:comparison}, we qualitative comparisons of the 3D human shape generation and human mesh recovery results. As visualized in Figure \ref{fig:comparison} (top), our approach generates more realistic poses than PoseNDF \cite{tiwari22posendf} and NRDF \cite{he24nrdf}. Figure \ref{fig:comparison} (middle and bottom) shows that Mamba-HMR produces reconstruction results with less distortion compared to METRO \cite{lin2021end-to-end}, which requires an additional upsampling process, and NLF \cite{sarandi2024nlf}, which requires a parametric model to obtain a dense mesh. Note that Mamba-HMR is able to project the resulting surfaces to a parametric pose representation by employing a method such as VPoser \cite{SMPL-X:2019}, but the quality of the reconstructions does not change significantly, with no large visual difference.

\subsection{Quantitative comparisons}

\noindent {\bf 3D Human Generation} In Table \ref{tab:generation}, we list the metric scores of MambaDiff3D and the baseline methods. MambaDiff3D outperforms both the parametric pose-based and non-parametric surface-based approaches. In fact, MambaDiff3D achieves state-of-the-art (SOTA) performance on the 1-NNA metric. The pose metric scores FID and APD further indicate that MambaDiff3D generates  diverse yet more realistic results than NRDF \cite{he24nrdf}, as depicted in Fig. \ref{fig:comparison}.

\noindent {\bf Whole-body human mesh recovery }  Table \ref{tab:hmr-wb2} presents the comparison of whole-body mesh recovery methods on UBody.  Mamba-HMR outperforms the SOTA parametric and non-parametric approaches \cite{smplerx, sun2024aios, multi-hmr2024, sarandi2024nlf}, including those pre-trained on a large-scale dataset.

\begin{figure}[t]
\begin{center}
 \includegraphics[width=0.95\linewidth]{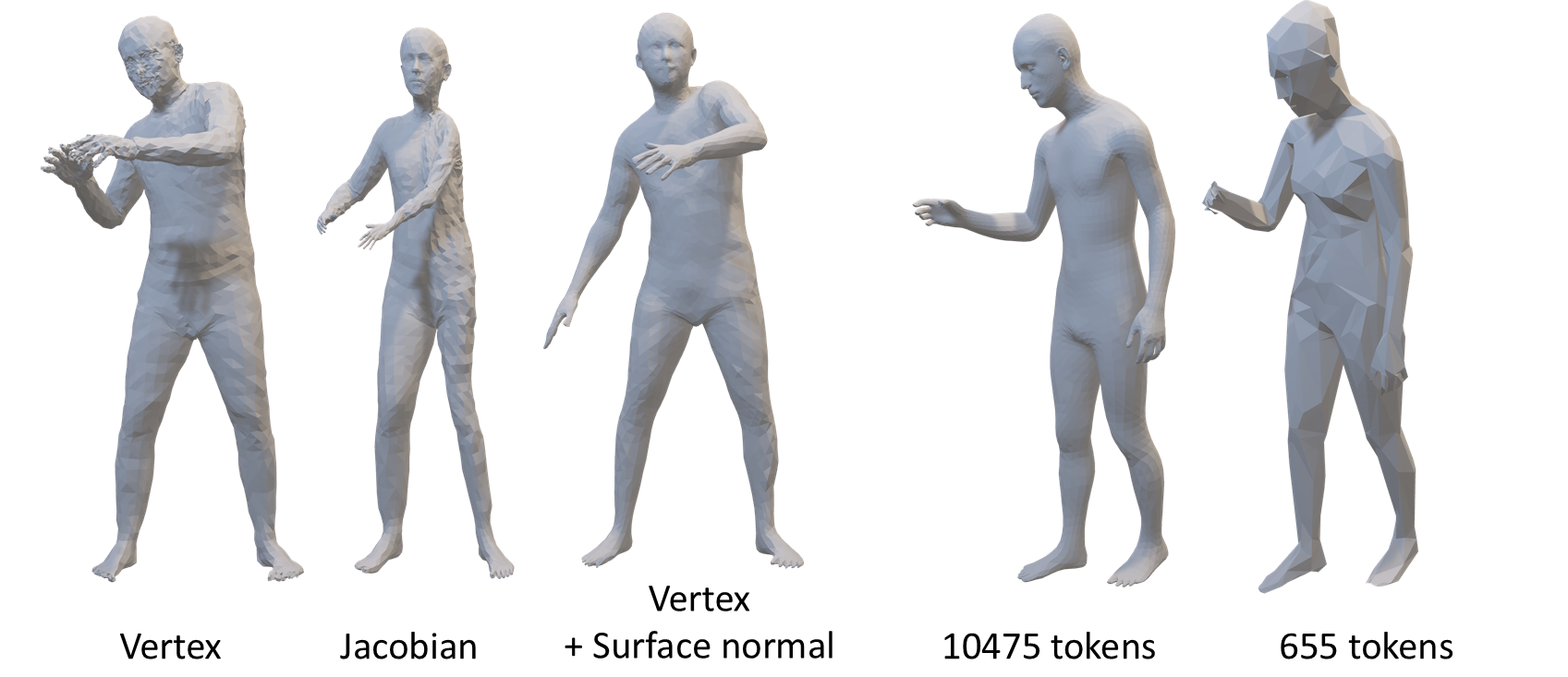}
 \vspace{-10pt}
 \caption{Comparisons of mesh representation in 3D generation. Left: Generation of vertices exhibits noise locally while Jacobians are prone to distortions globally. In contrast, our approach utilizing surface normals can preserve shape structure and achieves smooth reconstruction. Right: Performing generation on a downsampled mesh cannot recover hand shapes and loses fingers. } 
\vspace{-10pt}
 \label{fig:representation}
\end{center}
\end{figure}

\subsection{Ablation studies} 
\label{sec:ablation}
\noindent{\bf Network block } Table \ref{tab:ablation} (left) presents the results of ablation studies on network blocks, where we replaced the Mamba block in each layer with MLP, GNN and transformer self-attention. As shown, transformer and Mamba blocks perform significantly better than MLPs and GNNs, which led to unsuccessful training and produced locally very noisy surface results (see Appendix). 

\noindent{\bf Serialization } In Table \ref{tab:ablation} (right), we present the ablation study on mesh vertex serialization approaches. When vertices were sorted with a random ordering, MeshMamba was unable to learn properly. Using the default ordering derived from the SMPL mesh connectivity solely (``SMPL connectivity" in Table \ref{tab:ablation} (right)), it leads to a worse 1-NNA score and visually noticeable large distortions. With a single serialization strategy derived from the body part IUV maps, the 1-NNA score improved. Combining two or more serialization strategies leads to better 1-NNA scores but increasing the number of strategies needs a longer inference time due to memory access via ``gather" operations. Based on these results, we empirically found that a combination of two strategies balances efficiency and quality.

\noindent{\bf Mesh representation }  In Fig. \ref{fig:representation} we compared our mesh representation  that combines vertices and surface normals for mesh generation against generation vertices and Jacobians \cite{10.1145/1015706.1015736, aigerman2022neural, yoo2024neuralpose}. As reported in the recent works \cite{yoo2024neuralpose,muralikrishnan2024temporalresidualjacobiansrigfree}, the vertex and Jacobian generation  approaches are prone to noise and distortions. 
In contrast, our approach utilizing surface normals can preserve shape structure and achieves smooth reconstruction. Furthermore, performing generation on a downsampled mesh as in \cite{dat2023, yoshiyasu2024} cannot recover hand shapes and loses fingers.

\begin{figure}[t]
\begin{center}
 \includegraphics[width=0.9\linewidth]{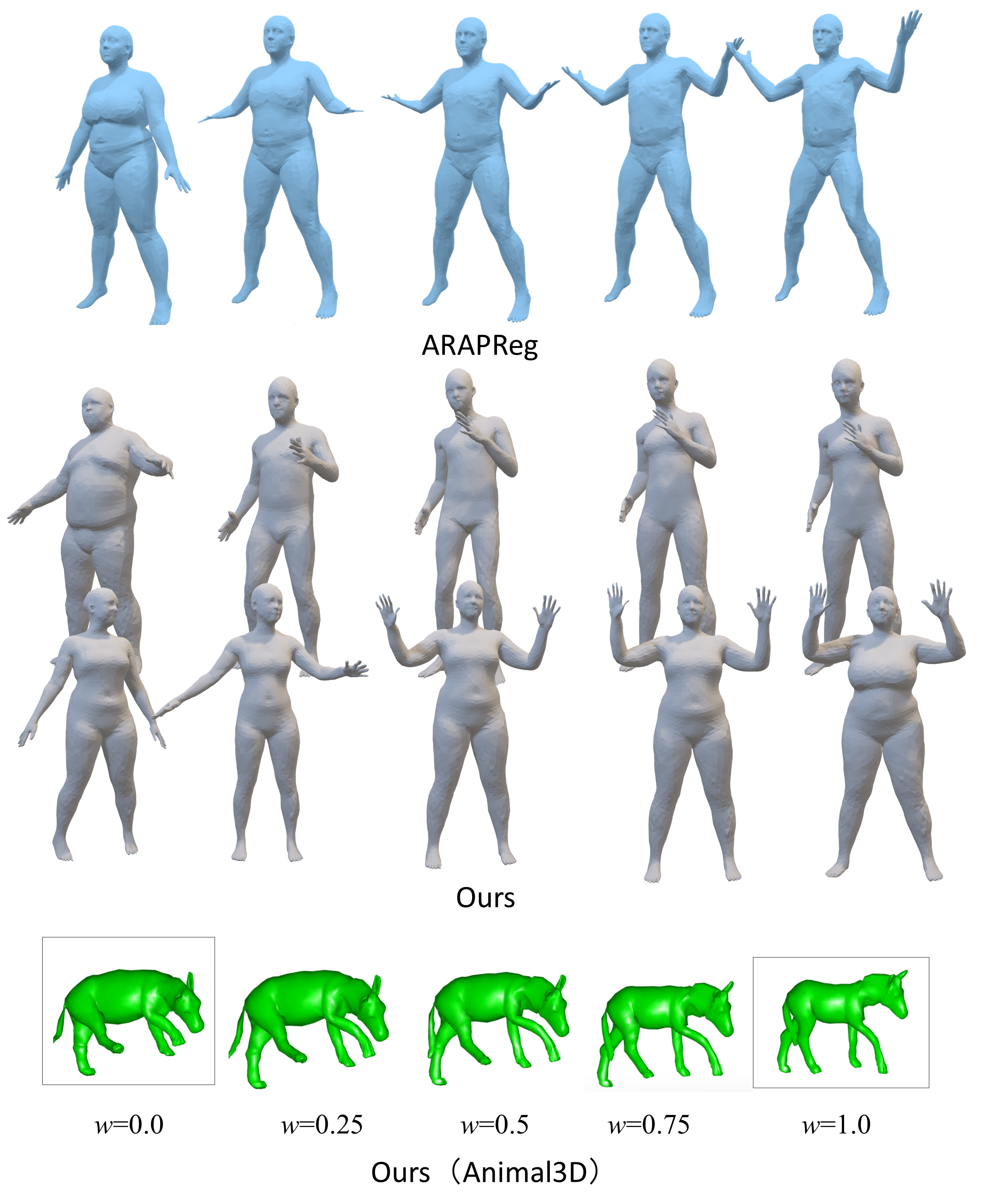}
 \vspace{-10pt}
 \caption{Shape interpolation. Compared to ARAPReg which enforces locally as-rigid-as possible constraints on mesh latent vectors, MambaDiff3D can faithfully preserves arm shapes while elbow bending. Also,  MambaDiff3D is capable of generating other mammals (3889 vertex tokens) by training on Animal3D datasets.  } 
 \label{fig:interpolation}
 \vspace{-10pt}
\end{center}
\end{figure}

\subsection{Shape interpolation}

Using MambaDiff3D, it is possible to perform shape interpolation by blending Gaussian noise with SLERP and sampling from the blended noise with the DDIM sampler. Compared to ARAPReg \cite{huang2021arapregasrigidaspossibleregularization} which enforces locally as-rigid-as possible constraints when constructing mesh latent vectors and performs interpolation based on them, MambaDiff3D can faithfully preserves arm shapes even when elbows are deeply bent during interpolation (Fig. \ref{fig:interpolation}). Additionally, MambaDiff3D is generalizable to other mammal models.  

\subsection{Limitations}

MeshMamba still has limitations that could be addressed in future research. First, it is limited to tight clothing with a fixed topology. We aim to tackle more challenging in-the-wild clothed human mesh recovery \cite{xiu2022icon}  by further increasing image and mesh resolution. Second, its generalization capability to new datasets not used in training is still limited, compared to approaches trained on diverse datasets \cite{smplerx}.

\section{Conclusion}

We presented MeshMamba, a neural network model for learning dense 3D articulated mesh models based on Mamba-SSMs. The key to effectively training MeshMamba lies in the serialization technique of mesh vertices, which leverages prior knowledge about the mesh structure encoded in a template mesh through its 3D coordinates or DensePose body part annotations. Building upon MeshMamba, we presented MambaDiff3D and Mamba-HMR for dense 3D human mesh generation and reconstruction. MambaDiff3D achieves state-of-the-art performance in the benchmark and, more importantly, it can generate 3D human meshes with clothing deformations and hand grasping poses. Mamba-HMR is the first Mamba-based approach to 3D human mesh recovery tasks, achieving competitive performance at a near real-time rate.

\section*{Acknowledgments } This work was supported by JSPS KAKENHI Grant Number 23K28116 and 22H00545 in Japan. This work was supported by  AIST policy-based budget project R\&D on Generative AI Foundation Models for the Physical Domain. We thank T. Murata for preparing baseline models and dataset.

{
    \small
    \bibliographystyle{ieeenat_fullname}
    \bibliography{egbib}

\begin{thebibliography}{97}
\providecommand{\natexlab}[1]{#1}
\providecommand{\url}[1]{\texttt{#1}}
\expandafter\ifx\csname urlstyle\endcsname\relax
  \providecommand{\doi}[1]{doi: #1}\else
  \providecommand{\doi}{doi: \begingroup \urlstyle{rm}\Url}\fi

\bibitem[Aigerman et~al.(2022)Aigerman, Gupta, Kim, Chaudhuri, Saito, and Groueix]{aigerman2022neural}
Noam Aigerman, Kunal Gupta, Vladimir~G Kim, Siddhartha Chaudhuri, Jun Saito, and Thibault Groueix.
\newblock Neural jacobian fields: Learning intrinsic mappings of arbitrary meshes.
\newblock \emph{SIGGRAPH}, 2022.

\bibitem[Aumentado-Armstrong et~al.(2019)Aumentado-Armstrong, Tsogkas, Jepson, and Dickinson]{9010824}
Tristan Aumentado-Armstrong, Stavros Tsogkas, Allan Jepson, and Sven Dickinson.
\newblock Geometric disentanglement for generative latent shape models.
\newblock In \emph{2019 IEEE/CVF International Conference on Computer Vision (ICCV)}, pages 8180--8189, 2019.

\bibitem[Bao et~al.(2023)Bao, Nie, Xue, Cao, Li, Su, and Zhu]{bao2022all}
Fan Bao, Shen Nie, Kaiwen Xue, Yue Cao, Chongxuan Li, Hang Su, and Jun Zhu.
\newblock All are worth words: A vit backbone for diffusion models.
\newblock In \emph{CVPR}, 2023.

\bibitem[Baradel* et~al.(2024)Baradel*, Armando, Galaaoui, Br{\'e}gier, Weinzaepfel, Rogez, and Lucas*]{multi-hmr2024}
Fabien Baradel*, Matthieu Armando, Salma Galaaoui, Romain Br{\'e}gier, Philippe Weinzaepfel, Gr{\'e}gory Rogez, and Thomas Lucas*.
\newblock Multi-hmr: Multi-person whole-body human mesh recovery in a single shot.
\newblock In \emph{ECCV}, 2024.

\bibitem[Biggs et~al.(2020)Biggs, Ehrhart, Joo, Graham, Vedaldi, and Novotny]{biggs2020multibodies}
Benjamin Biggs, S{\'{e}}bastien Ehrhart, Hanbyul Joo, Benjamin Graham, Andrea Vedaldi, and David Novotny.
\newblock {3D} multibodies: Fitting sets of plausible {3D} models to ambiguous image data.
\newblock In \emph{NeurIPS}, 2020.

\bibitem[Black et~al.(2023)Black, Patel, Tesch, and Yang]{bedlam}
Michael~J Black, Priyanka Patel, Joachim Tesch, and Jinlong Yang.
\newblock Bedlam: A synthetic dataset of bodies exhibiting detailed lifelike animated motion.
\newblock In \emph{CVPR}, pages 8726--8737, 2023.

\bibitem[Bogo et~al.(2016)Bogo, Kanazawa, Lassner, Gehler, Romero, and Black]{bogo2016keep}
Federica Bogo, Angjoo Kanazawa, Christoph Lassner, Peter Gehler, Javier Romero, and Michael~J Black.
\newblock Keep it smpl: Automatic estimation of 3d human pose and shape from a single image.
\newblock In \emph{ECCV}, pages 561--578. Springer, 2016.

\bibitem[Bogo et~al.(2017)Bogo, Romero, Pons-Moll, and Black]{dfaust2017}
Federica Bogo, Javier Romero, Gerard Pons-Moll, and Michael~J. Black.
\newblock Dynamic {FAUST}: {R}egistering human bodies in motion.
\newblock In \emph{IEEE Conf. on Computer Vision and Pattern Recognition (CVPR)}, 2017.

\bibitem[Botsch et~al.(2006)Botsch, Sumner, Pauly, and Gross]{Botsch2006DeformationTF}
Mario Botsch, Robert~W. Sumner, Mark Pauly, and Markus Gross.
\newblock Deformation transfer for detail-preserving surface editing.
\newblock 2006.

\bibitem[Cai et~al.(2023)Cai, Yin, Zeng, Wei, Sun, Wang, Pang, Mei, Zhang, Zhang, et~al.]{smplerx}
Zhongang Cai, Wanqi Yin, Ailing Zeng, Chen Wei, Qingping Sun, Yanjun Wang, Hui~En Pang, Haiyi Mei, Mingyuan Zhang, Lei Zhang, et~al.
\newblock Smpler-x: Scaling up expressive human pose and shape estimation.
\newblock In \emph{NeurIPs}, 2023.

\bibitem[Chen et~al.(2021)Chen, Tang, Shi, Peng, Sebe, and Zhao]{iepgan2021}
Haoyu Chen, Hao Tang, Henglin Shi, Wei Peng, Nicu Sebe, and Guoying Zhao.
\newblock Intrinsic-extrinsic preserved gans for unsupervised 3d pose transfer.
\newblock In \emph{2021 IEEE/CVF International Conference on Computer Vision (ICCV)}, pages 8610--8619, 2021.

\bibitem[Cheng et~al.(2019)Cheng, Bronstein, Zhou, Kotsia, Pantic, and Zafeiriou]{DBLP:journals/corr/abs-1903-10384}
Shiyang Cheng, Michael~M. Bronstein, Yuxiang Zhou, Irene Kotsia, Maja Pantic, and Stefanos Zafeiriou.
\newblock Meshgan: Non-linear 3d morphable models of faces.
\newblock \emph{CoRR}, abs/1903.10384, 2019.

\bibitem[Cho and Kim(2023)]{cho2023generative}
Hanbyel Cho and Junmo Kim.
\newblock Generative approach for probabilistic human mesh recovery using diffusion models, 2023.

\bibitem[Cho et~al.(2022)Cho, Youwang, and Oh]{cho_arxiv.2207.13820}
Junhyeong Cho, Kim Youwang, and Tae-Hyun Oh.
\newblock Cross-attention of disentangled modalities for 3d human mesh recovery with transformers.
\newblock In \emph{ECCV}, 2022.

\bibitem[Choi et~al.(2020)Choi, Moon, and Lee]{Choi_2020_ECCV_Pose2Mesh}
Hongsuk Choi, Gyeongsik Moon, and Kyoung~Mu Lee.
\newblock Pose2mesh: Graph convolutional network for 3d human pose and mesh recovery from a 2d human pose.
\newblock In \emph{ECCV}, 2020.

\bibitem[Cosmo et~al.(2020)Cosmo, Norelli, Halimi, Kimmel, and Rodol\`{a}]{LIMP2020}
Luca Cosmo, Antonio Norelli, Oshri Halimi, Ron Kimmel, and Emanuele Rodol\`{a}.
\newblock Limp: Learning latent shape representations with metric preservation priors.
\newblock In \emph{ECCV}, page 19–35, 2020.

\bibitem[Davydov et~al.(2022)Davydov, Remizova, Constantin, Honari, Salzmann, and Fua]{9879900}
A. Davydov, A. Remizova, V. Constantin, S. Honari, M. Salzmann, and P. Fua.
\newblock Adversarial parametric pose prior.
\newblock In \emph{2022 IEEE/CVF Conference on Computer Vision and Pattern Recognition (CVPR)}, pages 10987--10995, 2022.

\bibitem[Georgakis et~al.(2020)Georgakis, Li, Karanam, Chen, Ko{\v{s}}eck{\'a}, and Wu]{hierachical2020}
Georgios Georgakis, Ren Li, Srikrishna Karanam, Terrence Chen, Jana Ko{\v{s}}eck{\'a}, and Ziyan Wu.
\newblock Hierarchical kinematic human mesh recovery.
\newblock In \emph{ECCV}, pages 768--784. Springer International Publishing, 2020.

\bibitem[Goel et~al.(2023)Goel, Pavlakos, Rajasegaran, Kanazawa*, and Malik*]{goel2023humans}
Shubham Goel, Georgios Pavlakos, Jathushan Rajasegaran, Angjoo Kanazawa*, and Jitendra Malik*.
\newblock Humans in 4{D}: Reconstructing and tracking humans with transformers.
\newblock In \emph{ICCV}, 2023.

\bibitem[Gong et~al.(2023)Gong, Foo, Fan, Ke, Rahmani, and Liu]{gong2023diffpose}
Jia Gong, Lin~Geng Foo, Zhipeng Fan, Qiuhong Ke, Hossein Rahmani, and Jun Liu.
\newblock Diffpose: Toward more reliable 3d pose estimation.
\newblock In \emph{Proceedings of the IEEE/CVF Conference on Computer Vision and Pattern Recognition (CVPR)}, 2023.

\bibitem[Gu and Dao(2024)]{gu2024mamba}
Albert Gu and Tri Dao.
\newblock Mamba: Linear-time sequence modeling with selective state spaces, 2024.

\bibitem[Gu et~al.(2022)Gu, Goel, and Ré]{gu2022efficiently}
Albert Gu, Karan Goel, and Christopher Ré.
\newblock Efficiently modeling long sequences with structured state spaces, 2022.

\bibitem[He et~al.(2024)He, Tiwari, Birdal, Lenssen, and Pons-Moll]{he24nrdf}
Yannan He, Garvita Tiwari, Tolga Birdal, Jan~Eric Lenssen, and Gerard Pons-Moll.
\newblock Nrdf: Neural riemannian distance fields for learning articulated pose priors.
\newblock In \emph{Conference on Computer Vision and Pattern Recognition ({CVPR})}, 2024.

\bibitem[Heidari et~al.(2024)Heidari, Ghorbani~Kolahi, Karimijafarbigloo, Azad, Bozorgpour, Hatami, Azad, Diba, Bagci, Merhof, et~al.]{heidari2024computation}
Moein Heidari, Sina Ghorbani~Kolahi, Sanaz Karimijafarbigloo, Bobby Azad, Afshin Bozorgpour, Soheila Hatami, Reza Azad, Ali Diba, Ulas Bagci, Dorit Merhof, et~al.
\newblock Computation-efficient era: A comprehensive survey of state space models in medical image analysis.
\newblock \emph{arXiv e-prints}, pages arXiv--2406, 2024.

\bibitem[Ho et~al.(2020)Ho, Jain, and Abbeel]{ho2020denoising}
Jonathan Ho, Ajay Jain, and Pieter Abbeel.
\newblock Denoising diffusion probabilistic models.
\newblock \emph{arXiv preprint arxiv:2006.11239}, 2020.

\bibitem[Huang et~al.(2021)Huang, Huang, Sun, Zhang, Jiang, and Bajaj]{huang2021arapregasrigidaspossibleregularization}
Qixing Huang, Xiangru Huang, Bo Sun, Zaiwei Zhang, Junfeng Jiang, and Chandrajit Bajaj.
\newblock Arapreg: An as-rigid-as possible regularization loss for learning deformable shape generators, 2021.

\bibitem[Ionescu et~al.(2014)Ionescu, Papava, Olaru, and Sminchisescu]{h36m_pami}
Catalin Ionescu, Dragos Papava, Vlad Olaru, and Cristian Sminchisescu.
\newblock Human3.6m: Large scale datasets and predictive methods for 3d human sensing in natural environments.
\newblock \emph{IEEE TPAMI}, 36\penalty0 (7):\penalty0 1325--1339, 2014.

\bibitem[Jiang et~al.(2020)Jiang, Zhang, Cai, and Zheng]{Jiang2020HumanBody}
Boyi Jiang, Juyong Zhang, Jianfei Cai, and Jianmin Zheng.
\newblock Disentangled human body embedding based on deep hierarchical neural network.
\newblock 2020.

\bibitem[Jin et~al.(2020)Jin, Xu, Xu, Wang, Liu, Qian, Ouyang, and Luo]{jin2020whole}
Sheng Jin, Lumin Xu, Jin Xu, Can Wang, Wentao Liu, Chen Qian, Wanli Ouyang, and Ping Luo.
\newblock Whole-body human pose estimation in the wild.
\newblock In \emph{ECCV}, 2020.

\bibitem[Kanazawa et~al.(2018)Kanazawa, Black, Jacobs, and Malik]{hmrKanazawa17}
Angjoo Kanazawa, Michael~J. Black, David~W. Jacobs, and Jitendra Malik.
\newblock End-to-end recovery of human shape and pose.
\newblock In \emph{CVPR}, 2018.

\bibitem[Kim et~al.(2023)Kim, Gwon, Park, Kwon, Um, and Kim]{PointHMR}
Jeonghwan Kim, Mi-Gyeong Gwon, Hyunwoo Park, Hyukmin Kwon, Gi-Mun Um, and Wonjun Kim.
\newblock {Sampling is Matter}: Point-guided 3d human mesh reconstruction.
\newblock In \emph{CVPR}, 2023.

\bibitem[Kingma and Gao(2023)]{NEURIPS2023_kingma}
Diederik Kingma and Ruiqi Gao.
\newblock Understanding diffusion objectives as the elbo with simple data augmentation.
\newblock In \emph{NeurIPS}, pages 65484--65516, 2023.

\bibitem[Kolotouros et~al.(2019)Kolotouros, Pavlakos, and Daniilidis]{kolotouros2019cmr}
Nikos Kolotouros, Georgios Pavlakos, and Kostas Daniilidis.
\newblock Convolutional mesh regression for single-image human shape reconstruction.
\newblock In \emph{CVPR}, 2019.

\bibitem[Kolotouros et~al.(2021)Kolotouros, Pavlakos, Jayaraman, and Daniilidis]{kolotouros2021prohmr}
Nikos Kolotouros, Georgios Pavlakos, Dinesh Jayaraman, and Kostas Daniilidis.
\newblock Probabilistic modeling for human mesh recovery.
\newblock In \emph{ICCV}, 2021.

\bibitem[Li et~al.(2023)Li, Zhuo, Zhang, Bo, and Chen]{li2023diffhand}
Lijun Li, Li'an Zhuo, Bang Zhang, Liefeng Bo, and Chen Chen.
\newblock Diffhand: End-to-end hand mesh reconstruction via diffusion models, 2023.

\bibitem[Liang et~al.(2024)Liang, Zhou, Xu, Zhu, Zou, Ye, Tan, and Bai]{liang2024pointmamba}
Dingkang Liang, Xin Zhou, Wei Xu, Xingkui Zhu, Zhikang Zou, Xiaoqing Ye, Xiao Tan, and Xiang Bai.
\newblock Pointmamba: A simple state space model for point cloud analysis.
\newblock \emph{arXiv preprint arXiv:2402.10739}, 2024.

\bibitem[Lin et~al.(2023)Lin, Zeng, Wang, Zhang, and Li]{osx}
Jing Lin, Ailing Zeng, Haoqian Wang, Lei Zhang, and Yu Li.
\newblock One-stage 3d whole-body mesh recovery with component aware transformer.
\newblock 2023.

\bibitem[Lin et~al.(2021{\natexlab{a}})Lin, Wang, and Liu]{lin2021-mesh-graphormer}
Kevin Lin, Lijuan Wang, and Zicheng Liu.
\newblock Mesh graphormer.
\newblock In \emph{ICCV}, 2021{\natexlab{a}}.

\bibitem[Lin et~al.(2021{\natexlab{b}})Lin, Wang, and Liu]{lin2021end-to-end}
Kevin Lin, Lijuan Wang, and Zicheng Liu.
\newblock End-to-end human pose and mesh reconstruction with transformers.
\newblock In \emph{CVPR}, 2021{\natexlab{b}}.

\bibitem[Lin et~al.(2014)Lin, Maire, Belongie, Bourdev, Girshick, Hays, Perona, Ramanan, Doll{\'{a}}r, and Zitnick]{LinMBHPRDZ14}
Tsung{-}Yi Lin, Michael Maire, Serge~J. Belongie, Lubomir~D. Bourdev, Ross~B. Girshick, James Hays, Pietro Perona, Deva Ramanan, Piotr Doll{\'{a}}r, and C.~Lawrence Zitnick.
\newblock Microsoft {COCO:} common objects in context.
\newblock \emph{CoRR}, abs/1405.0312, 2014.

\bibitem[Lin Geng~Foo(2023)]{dat2023}
Hossein Rahmani Jun~Liu Lin Geng~Foo, Jia~Gong.
\newblock Distribution-aligned diffusion for human mesh recovery.
\newblock In \emph{ICCV}, 2023.

\bibitem[Liu et~al.(2024)Liu, Yu, Wang, Zheng, Deng, Ye, and Wang]{liu2024point}
Jiuming Liu, Ruiji Yu, Yian Wang, Yu Zheng, Tianchen Deng, Weicai Ye, and Hesheng Wang.
\newblock Point mamba: A novel point cloud backbone based on state space model with octree-based ordering strategy, 2024.

\bibitem[Liu et~al.(2023)Liu, Yang, Gu, Guo, and Yang]{10161247}
Yuxuan Liu, Jianxin Yang, Xiao Gu, Yao Guo, and Guang-Zhong Yang.
\newblock Egohmr: Egocentric human mesh recovery via hierarchical latent diffusion model.
\newblock In \emph{2023 IEEE International Conference on Robotics and Automation (ICRA)}, pages 9807--9813, 2023.

\bibitem[Loper et~al.(2015)Loper, Mahmood, Romero, Pons-Moll, and Black]{SMPL:2015}
Matthew Loper, Naureen Mahmood, Javier Romero, Gerard Pons-Moll, and Michael~J. Black.
\newblock {SMPL}: A skinned multi-person linear model.
\newblock \emph{ACM Trans. Graphics (Proc. SIGGRAPH Asia)}, 34\penalty0 (6):\penalty0 248:1--248:16, 2015.

\bibitem[Lu et~al.(2024)Lu, Lin, Dou, Zeng, Deng, Zhang, and Wang]{lu2024dposer}
Junzhe Lu, Jing Lin, Hongkun Dou, Ailing Zeng, Yue Deng, Yulun Zhang, and Haoqian Wang.
\newblock Dposer: Diffusion model as robust 3d human pose prior, 2024.

\bibitem[Luo and Hu(2021)]{luo2021diffusion}
Shitong Luo and Wei Hu.
\newblock Diffusion probabilistic models for 3d point cloud generation.
\newblock In \emph{Proceedings of the IEEE/CVF Conference on Computer Vision and Pattern Recognition (CVPR)}, 2021.

\bibitem[Ma et~al.(2020)Ma, Yang, Ranjan, Pujades, Pons-Moll, Tang, and Black]{ma2020cape}
Qianli Ma, Jinlong Yang, Anurag Ranjan, Sergi Pujades, Gerard Pons-Moll, Siyu Tang, and Michael~J. Black.
\newblock Learning to dress 3d people in generative clothing.
\newblock In \emph{Computer Vision and Pattern Recognition (CVPR)}, 2020.

\bibitem[Ma et~al.(2021)Ma, Yang, Tang, and Black]{POP:ICCV:2021}
Qianli Ma, Jinlong Yang, Siyu Tang, and Michael~J. Black.
\newblock The power of points for modeling humans in clothing.
\newblock In \emph{Proceedings of the IEEE/CVF International Conference on Computer Vision (ICCV)}, pages 10974--10984, 2021.

\bibitem[Ma et~al.(2023)Ma, Su, Wang, Zhu, and Wang]{Ma_2023_CVPR}
Xiaoxuan Ma, Jiajun Su, Chunyu Wang, Wentao Zhu, and Yizhou Wang.
\newblock 3d human mesh estimation from virtual markers.
\newblock In \emph{Proceedings of the IEEE/CVF Conference on Computer Vision and Pattern Recognition (CVPR)}, pages 534--543, 2023.

\bibitem[Mahmood et~al.(2019)Mahmood, Ghorbani, F.~Troje, Pons-Moll, and Black]{AMASS:2019}
Naureen Mahmood, Nima Ghorbani, Nikolaus F.~Troje, Gerard Pons-Moll, and Michael~J. Black.
\newblock Amass: Archive of motion capture as surface shapes.
\newblock In \emph{The IEEE International Conference on Computer Vision (ICCV)}, 2019.

\bibitem[Muralikrishnan et~al.(2024)Muralikrishnan, Dutt, Chaudhuri, Aigerman, Kim, Fisher, and Mitra]{muralikrishnan2024temporalresidualjacobiansrigfree}
Sanjeev Muralikrishnan, Niladri~Shekhar Dutt, Siddhartha Chaudhuri, Noam Aigerman, Vladimir Kim, Matthew Fisher, and Niloy~J. Mitra.
\newblock Temporal residual jacobians for rig-free motion transfer, 2024.

\bibitem[Nehab et~al.(2005)Nehab, Rusinkiewicz, Davis, and Ramamoorthi]{nehab2005}
Diego Nehab, Szymon Rusinkiewicz, James Davis, and Ravi Ramamoorthi.
\newblock Efficiently combining positions and normals for precise 3d geometry.
\newblock \emph{ACM Trans. Graph.}, 24\penalty0 (3):\penalty0 536–543, 2005.

\bibitem[Patel et~al.(2021)Patel, Huang, Tesch, Hoffmann, Tripathi, and Black]{agora2021}
Priyanka Patel, Chun-Hao~P. Huang, Joachim Tesch, David~T. Hoffmann, Shashank Tripathi, and Michael~J. Black.
\newblock {AGORA}: Avatars in geography optimized for regression analysis.
\newblock In \emph{Proceedings IEEE/CVF Conf.~on Computer Vision and Pattern Recognition ({CVPR})}, 2021.

\bibitem[Patro and Agneeswaran(2024)]{patro2024mamba}
Badri~Narayana Patro and Vijay~Srinivas Agneeswaran.
\newblock Mamba-360: Survey of state space models as transformer alternative for long sequence modelling: Methods, applications, and challenges.
\newblock \emph{arXiv preprint arXiv:2404.16112}, 2024.

\bibitem[Pavlakos et~al.(2019)Pavlakos, Choutas, Ghorbani, Bolkart, Osman, Tzionas, and Black]{SMPL-X:2019}
Georgios Pavlakos, Vasileios Choutas, Nima Ghorbani, Timo Bolkart, Ahmed A.~A. Osman, Dimitrios Tzionas, and Michael~J. Black.
\newblock Expressive body capture: 3d hands, face, and body from a single image.
\newblock In \emph{Proceedings IEEE Conf. on Computer Vision and Pattern Recognition (CVPR)}, 2019.

\bibitem[Pavlakos et~al.(2020)Pavlakos, Choutas, Bolkart, Tzionas, Black, Choutas, Pavlakos, Bolkart, Tzionas, and Black]{Pavlakos2020expose}
Georgios Pavlakos, Vasileios Choutas, Timo Bolkart, Dimitrios Tzionas, Michael~J. Black, Vasileios Choutas, Georgios Pavlakos, Timo Bolkart, Dimitrios Tzionas, and Michael~J. Black.
\newblock Monocular expressive body regression through body-driven attention.
\newblock \emph{ECCV}, 2020.

\bibitem[Ranjan et~al.(2018)Ranjan, Bolkart, Sanyal, and Black]{COMA:ECCV18}
Anurag Ranjan, Timo Bolkart, Soubhik Sanyal, and Michael~J. Black.
\newblock Generating {3D} faces using convolutional mesh autoencoders.
\newblock In \emph{European Conference on Computer Vision (ECCV)}, pages 725--741, 2018.

\bibitem[Riza et~al.(2018)Riza, Natalia, and Iasonas]{Guler2018DensePose}
Guler Riza, Neverova Natalia, and Kokkinos Iasonas.
\newblock Densepose: Dense human pose estimation in the wild.
\newblock \emph{arXiv}, 2018.

\bibitem[Rombach et~al.(2022)Rombach, Blattmann, Lorenz, Esser, and Ommer]{rombach_2022_CVPR}
Robin Rombach, Andreas Blattmann, Dominik Lorenz, Patrick Esser, and Bj\"orn Ommer.
\newblock High-resolution image synthesis with latent diffusion models.
\newblock In \emph{Proceedings of the IEEE/CVF Conference on Computer Vision and Pattern Recognition (CVPR)}, pages 10684--10695, 2022.

\bibitem[Rong et~al.(2021)Rong, Shiratori, and Joo]{frankmocap}
Yu Rong, Takaaki Shiratori, and Hanbyul Joo.
\newblock Frankmocap: A monocular 3d whole-body pose estimation system via regression and integration.
\newblock In \emph{ICCVW}, 2021.

\bibitem[Rueegg et~al.(2022)Rueegg, Zuffi, Schindler, and Black]{BARC:2022}
Nadine Rueegg, Silvia Zuffi, Konrad Schindler, and Michael~J. Black.
\newblock Barc: Learning to regress 3d dog shape from images by exploiting breed information.
\newblock In \emph{Proceedings IEEE Conf. on Computer Vision and Pattern Recognition (CVPR)}, 2022.

\bibitem[Salimans and Ho(2022)]{salimans2022progressive}
Tim Salimans and Jonathan Ho.
\newblock Progressive distillation for fast sampling of diffusion models, 2022.

\bibitem[S\'ar\'andi and Pons-Moll(2024)]{sarandi2024nlf}
Istv\'an S\'ar\'andi and Gerard Pons-Moll.
\newblock Neural localizer fields for continuous 3d human pose and shape estimation.
\newblock 2024.

\bibitem[Shan et~al.(2023)Shan, Liu, Zhang, Wang, Han, Wang, Ma, and Gao]{shan2023diffusion}
Wenkang Shan, Zhenhua Liu, Xinfeng Zhang, Zhao Wang, Kai Han, Shanshe Wang, Siwei Ma, and Wen Gao.
\newblock Diffusion-based 3d human pose estimation with multi-hypothesis aggregation.
\newblock \emph{arXiv preprint arXiv:2303.11579}, 2023.

\bibitem[Song et~al.(2020)Song, Meng, and Ermon]{song2020denoising}
Jiaming Song, Chenlin Meng, and Stefano Ermon.
\newblock Denoising diffusion implicit models.
\newblock \emph{arXiv:2010.02502}, 2020.

\bibitem[Stathopoulos et~al.(2024)Stathopoulos, Han, and Metaxas]{stathopoulos2024scoreguided}
Anastasis Stathopoulos, Ligong Han, and Dimitris Metaxas.
\newblock Score-guided diffusion for 3d human recovery, 2024.

\bibitem[Sumner and Popovi\'{c}(2004)]{10.1145/1015706.1015736}
Robert~W. Sumner and Jovan Popovi\'{c}.
\newblock Deformation transfer for triangle meshes.
\newblock \emph{SIGGRAPH}, 23\penalty0 (3), 2004.

\bibitem[Sun et~al.(2019)Sun, Xiao, Liu, and Wang]{sun2019deep}
Ke Sun, Bin Xiao, Dong Liu, and Jingdong Wang.
\newblock Deep high-resolution representation learning for human pose estimation.
\newblock In \emph{CVPR}, 2019.

\bibitem[Sun et~al.(2024)Sun, Wang, Zeng, Yin, Wei, Wang, Mei, Leung, Liu, Yang, and Cai]{sun2024aios}
Qingping Sun, Yanjun Wang, Ailing Zeng, Wanqi Yin, Chen Wei, Wenjia Wang, Haiyi Mei, Chi~Sing Leung, Ziwei Liu, Lei Yang, and Zhongang Cai.
\newblock Aios: All-in-one-stage expressive human pose and shape estimation.
\newblock In \emph{CVPR}, 2024.

\bibitem[Taheri et~al.(2020)Taheri, Ghorbani, Black, and Tzionas]{GRAB:2020}
Omid Taheri, Nima Ghorbani, Michael~J. Black, and Dimitrios Tzionas.
\newblock {GRAB}: A dataset of whole-body human grasping of objects.
\newblock In \emph{ECCV}, 2020.

\bibitem[Tan et~al.(2018)Tan, Gao, Lai, and Xia]{8578710}
Qingyang Tan, Lin Gao, Yu-Kun Lai, and Shihong Xia.
\newblock Variational autoencoders for deforming 3d mesh models.
\newblock In \emph{2018 IEEE/CVF Conference on Computer Vision and Pattern Recognition}, pages 5841--5850, 2018.

\bibitem[Tay et~al.(2022)Tay, Dehghani, Bahri, and Metzler]{10.1145/3530811}
Yi Tay, Mostafa Dehghani, Dara Bahri, and Donald Metzler.
\newblock Efficient transformers: A survey.
\newblock \emph{ACM Comput. Surv.}, 2022.

\bibitem[Tian et~al.(2022)Tian, Zhang, Liu, and Wang]{tian2022hmrsurvey}
Yating Tian, Hongwen Zhang, Yebin Liu, and Limin Wang.
\newblock Recovering 3d human mesh from monocular images: A survey.
\newblock \emph{arXiv preprint arXiv:2203.01923}, 2022.

\bibitem[Tiwari et~al.(2022)Tiwari, Antic, Lenssen, Sarafianos, Tung, and Pons-Moll]{tiwari22posendf}
Garvita Tiwari, Dimitrije Antic, Jan~Eric Lenssen, Nikolaos Sarafianos, Tony Tung, and Gerard Pons-Moll.
\newblock Pose-ndf: Modeling human pose manifolds with neural distance fields.
\newblock In \emph{ECCV}, 2022.

\bibitem[Varol et~al.(2017)Varol, Romero, Martin, Mahmood, Black, Laptev, and Schmid]{varol17_surreal}
G{\"u}l Varol, Javier Romero, Xavier Martin, Naureen Mahmood, Michael~J. Black, Ivan Laptev, and Cordelia Schmid.
\newblock Learning from synthetic humans.
\newblock In \emph{CVPR}, 2017.

\bibitem[Wang et~al.(2024)Wang, Wang, Ding, Li, Wu, Rong, Kong, Huang, Li, Yang, Wang, Jiang, Li, Wang, Tian, and Tang]{wang2024statespacemodelnewgeneration}
Xiao Wang, Shiao Wang, Yuhe Ding, Yuehang Li, Wentao Wu, Yao Rong, Weizhe Kong, Ju Huang, Shihao Li, Haoxiang Yang, Ziwen Wang, Bo Jiang, Chenglong Li, Yaowei Wang, Yonghong Tian, and Jin Tang.
\newblock State space model for new-generation network alternative to transformers: A survey, 2024.

\bibitem[Weber et~al.(2007)Weber, Sorkine, Lipman, and Gotsman]{weber2007}
Ofir Weber, Olga Sorkine, Yaron Lipman, and Craig Gotsman.
\newblock Context-aware skeletal shape deformation.
\newblock \emph{Computer Graphics Forum}, 26\penalty0 (3):\penalty0 265--274, 2007.

\bibitem[Xiaokun~Sun(2023)]{SemanticHuman}
Xiongzheng Li Jinsong Zhang Yu-Kun Lai Jingyu Yang Kun~Li Xiaokun~Sun, Qiao~Feng.
\newblock Learning semantic-aware disentangled representation for flexible 3d human body editing.
\newblock In \emph{CVPR}, 2023.

\bibitem[Xiu et~al.(2022)Xiu, Yang, Tzionas, and Black]{xiu2022icon}
Yuliang Xiu, Jinlong Yang, Dimitrios Tzionas, and Michael~J. Black.
\newblock {ICON}: {I}mplicit {C}lothed humans {O}btained from {N}ormals.
\newblock In \emph{Proceedings of the IEEE/CVF Conference on Computer Vision and Pattern Recognition (CVPR)}, pages 13296--13306, 2022.

\bibitem[Xu et~al.(2020)Xu, Bazavan, Zanfir, Freeman, Sukthankar, and Sminchisescu]{xu2020ghum}
Hongyi Xu, Eduard~Gabriel Bazavan, Andrei Zanfir, William~T Freeman, Rahul Sukthankar, and Cristian Sminchisescu.
\newblock Ghum \& ghuml: Generative 3d human shape and articulated pose models.
\newblock In \emph{Proceedings of the IEEE/CVF Conference on Computer Vision and Pattern Recognition}, pages 6184--6193, 2020.

\bibitem[Xu et~al.(2023)Xu, Zhang, Peng, Ma, Jesslen, Ji, Hu, Zhang, Liu, Wang, et~al.]{xu2023animal3d}
Jiacong Xu, Yi Zhang, Jiawei Peng, Wufei Ma, Artur Jesslen, Pengliang Ji, Qixin Hu, Jiehua Zhang, Qihao Liu, Jiahao Wang, et~al.
\newblock Animal3d: A comprehensive dataset of 3d animal pose and shape.
\newblock \emph{arXiv preprint arXiv:2308.11737}, 2023.

\bibitem[Yang et~al.(2019)Yang, Huang, Hao, Liu, Belongie, and Hariharan]{pointflow}
Guandao Yang, Xun Huang, Zekun Hao, Ming-Yu Liu, Serge Belongie, and Bharath Hariharan.
\newblock Pointflow: 3d point cloud generation with continuous normalizing flows.
\newblock \emph{arXiv}, 2019.

\bibitem[Yoo et~al.(2024)Yoo, Koo, Yeo, and Sung]{yoo2024neuralpose}
Seungwoo Yoo, Juil Koo, Kyeongmin Yeo, and Minhyuk Sung.
\newblock {Neural Pose Representation Learning for Generating and Transferring Non-Rigid Object Poses}.
\newblock In \emph{NeurIPS}, 2024.

\bibitem[Yoshiyasu(2023)]{yoshiyasu2023-deformer}
Yusuke Yoshiyasu.
\newblock Deformable mesh transformer for 3d human mesh recovery.
\newblock In \emph{CVPR}, pages 17006--17015, 2023.

\bibitem[Yoshiyasu and Sun(2024)]{yoshiyasu2024}
Yusuke Yoshiyasu and Leyuan Sun.
\newblock Diffsurf: A transformer-based diffusion model for generating and reconstructing 3d surfaces in pose.
\newblock In \emph{ECCV}, 2024.

\bibitem[You et~al.(2023)You, Liu, Li, Li, Wang, and Ding]{10096870}
Yingxuan You, Hong Liu, Xia Li, Wenhao Li, Ti Wang, and Runwei Ding.
\newblock Gator: Graph-aware transformer with motion-disentangled regression for human mesh recovery from a 2d pose.
\newblock In \emph{ICASSP 2023 - 2023 IEEE International Conference on Acoustics, Speech and Signal Processing (ICASSP)}, pages 1--5, 2023.

\bibitem[Yuan et~al.(2020)Yuan, Lai, Yang, Duan, Fu, and Gao]{yuan2020mesh}
Yu-Jie Yuan, Yu-Kun Lai, Jie Yang, Qi Duan, Hongbo Fu, and Lin Gao.
\newblock Mesh variational autoencoders with edge contraction pooling.
\newblock In \emph{CVPRW}, pages 274--275, 2020.

\bibitem[Zanfir et~al.(2020)Zanfir, Bazavan, Xu, Freeman, Sukthankar, and Sminchisescu]{zanfir2020weakly}
Andrei Zanfir, Eduard~Gabriel Bazavan, Hongyi Xu, William~T. Freeman, Rahul Sukthankar, and Cristian Sminchisescu.
\newblock Weakly supervised 3d human pose and shape reconstruction with normalizing flows.
\newblock In \emph{Computer Vision -- ECCV 2020}, pages 465--481, 2020.

\bibitem[Zeng et~al.(2022)Zeng, Vahdat, Williams, Gojcic, Litany, Fidler, and Kreis]{zeng2022lion}
Xiaohui Zeng, Arash Vahdat, Francis Williams, Zan Gojcic, Or Litany, Sanja Fidler, and Karsten Kreis.
\newblock Lion: Latent point diffusion models for 3d shape generation.
\newblock In \emph{Advances in Neural Information Processing Systems (NeurIPS)}, 2022.

\bibitem[Zhang et~al.(2021)Zhang, Tian, Zhou, Ouyang, Liu, Wang, and Sun]{pymaf2021}
Hongwen Zhang, Yating Tian, Xinchi Zhou, Wanli Ouyang, Yebin Liu, Limin Wang, and Zhenan Sun.
\newblock Pymaf: 3d human pose and shape regression with pyramidal mesh alignment feedback loop.
\newblock In \emph{ICCV}, 2021.

\bibitem[Zhang et~al.(2022)Zhang, Tian, Zhang, Li, An, Sun, and Liu]{pymafx2022}
Hongwen Zhang, Yating Tian, Yuxiang Zhang, Mengcheng Li, Liang An, Zhenan Sun, and Yebin Liu.
\newblock Pymaf-x: Towards well-aligned full-body model regression from monocular images.
\newblock \emph{arXiv preprint arXiv:2207.06400}, 2022.

\bibitem[Zhang et~al.(2024{\natexlab{a}})Zhang, Bhatnagar, Xu, Winkler, Kadlecek, Tang, and Bogo]{zhang2024rohm}
Siwei Zhang, Bharat~Lal Bhatnagar, Yuanlu Xu, Alexander Winkler, Petr Kadlecek, Siyu Tang, and Federica Bogo.
\newblock Rohm: Robust human motion reconstruction via diffusion, 2024{\natexlab{a}}.

\bibitem[Zhang et~al.(2024{\natexlab{b}})Zhang, Li, Yuan, Ji, and Yan]{zhang2024point}
Tao Zhang, Xiangtai Li, Haobo Yuan, Shunping Ji, and Shuicheng Yan.
\newblock Point cloud mamba: Point cloud learning via state space model.
\newblock \emph{arXiv preprint arXiv:2403.00762}, 2024{\natexlab{b}}.

\bibitem[Zheng et~al.(2023)Zheng, Yifan, Wetzstein, Black, and Hilliges]{Zheng2023pointavatar}
Yufeng Zheng, Wang Yifan, Gordon Wetzstein, Michael~J. Black, and Otmar Hilliges.
\newblock Pointavatar: Deformable point-based head avatars from videos.
\newblock In \emph{Proceedings of the IEEE/CVF Conference on Computer Vision and Pattern Recognition (CVPR)}, 2023.

\bibitem[Zhengcong~Fei(2024)]{FeiDiS2024}
Changqian Yu Jusnshi~Huang Zhengcong~Fei, Mingyuan~Fan.
\newblock Scalable diffusion models with state space backbone.
\newblock \emph{arXiv preprint}, 2024.

\bibitem[Zhou et~al.(2020)Zhou, Bhatnagar, and Pons-Moll]{zhou20unsupervised}
Keyang Zhou, Bharat~Lal Bhatnagar, and Gerard Pons-Moll.
\newblock Unsupervised shape and pose disentanglement for 3d meshes.
\newblock In \emph{European Conference on Computer Vision (ECCV)}, 2020.

\bibitem[Zhu et~al.(2024)Zhu, Liao, Zhang, Wang, Liu, and Wang]{vim}
Lianghui Zhu, Bencheng Liao, Qian Zhang, Xinlong Wang, Wenyu Liu, and Xinggang Wang.
\newblock Vision mamba: Efficient visual representation learning with bidirectional state space model.
\newblock \emph{arXiv preprint arXiv:2401.09417}, 2024.

\end{thebibliography}
}

\end{document}